# Training Neural Networks to Approach the Optimum Bayes Estimator in Dense Multi-Emitter Localization

YI SUN*, MONA SHARIFI, AND MUZNA YUMMAN

*Electrical Engineering Department*
*Applied and Embodied AI Lab*
*The City College of City University of New York*
*New York, NY 10031, USA*
**Email: ysun@ccny.cuny.edu*
*ORCID: 0000-0002-2527-8311*

**Abstract:** Multi-emitter localization from a single frame sets the density limit of single molecule localization microscopy (SMLM), i.e., the accuracy falls as the emitters overlap, whereas at severe overlap the Cramér-Rao bound diverges and the best unbiased estimator loses accuracy without limit. The optimum estimator in the mean square sense is the Bayes estimator that minimizes the risk over a prior of emitter configurations, whereas for two or more emitters it has no closed form and its evaluation needs the point spread function, the emitter intensity and the per-pixel noise, so a direct computation requires the full system model. We train neural networks on synthesized frames to approach this estimator. Five networks that span learnable activations, Kolmogorov-Arnold edges, convolution, and self attention are trained with a permutation-matched loss whose Bayes estimator is exactly the reported metric, then compared against the unbiased Gaussian information-achieving estimator for a frame (UGIA-F) and the expectation-maximization global maximum likelihood estimator (EM-GML). The networks estimate from a frame alone, whereas EM-GML uses the system parameters and UGIA-F uses the system parameters and the true positions, so among the three only UGIA-F is an oracle and the networks and EM-GML are practically useful. Simulation results show that on in-distribution frames every network stays far below UGIA-F in mean square error where the emitters overlap and approaches EM-GML, e.g., the vision Transformer attains an average error of 31.6 nm against 30.3 nm for EM-GML and 172 nm for UGIA-F at a density of 56 emitters per μm$^2$. The networks remain bounded where the oracle UGIA-F diverges, whereas they reach the accuracy of EM-GML, which confirms the hypothesis in a proof of concept on a field of view (FOV) of $700 \times 700$ nm$^2$. The result justifies the future work on training neural networks to approach the optimum Bayes estimator on the large-size frames to achieve high-throughput large-FOV super spatiotemporal resolution SMLM.

## 1. Introduction

Single molecule localization microscopy (SMLM) [1-4] forms a super-resolution image by estimating the positions of individual fluorescent emitters from a sequence of diffraction limited frames. The resolution of the reconstruction and the temporal resolution, defined by the time needed to acquire it, are set against each other by the density of active emitters per frame, i.e., a sparse frame carries well separated emitters to be localized easily but many frames at a low temporal resolution are needed, whereas a dense frame shortens the acquisition, thus improving the temporal resolution, but places several overlapping emitters in a single point spread function (PSF). The estimation of severely overlapping emitters from one frame is therefore critical to achieve both super spatial and temporal resolutions, and it is the problem studied in this paper.

The accuracy of localization has long been analyzed through the Cramér-Rao bound (CRB), i.e., the lower bound on the covariance of any unbiased estimator, which for a single emitter gives the familiar dependence of the accuracy on the photon count and the background [5-7]. The accuracy depends on the PSF model, i.e., the Gaussian model in two dimensions [1] and the engineered functions that encode depth in three dimensions [9-12], and on the information content of the frame that sets the achievable segmentation and signal to noise ratio [13, 14]. The bound extends to several emitters through the Fisher information matrix of the multi-emitter frame [15-17], and the unbiased Gaussian information achieving estimator for a frame (UGIA-F) realizes it as an oracle benchmark that attains the CRB on every frame [18, 19]. In counterpart, the global maximum likelihood (GML) estimator [19] is a practical estimator, which is approached by multi-start expectation maximization (EM) and is denoted EM-GML [19-21]. Both benchmark estimators have limits that matter exactly where the emitters overlap. The CRB diverges when two emitters merge, so the accuracy attainable under the unbiasedness constraint is unbounded and UGIA-F degrades without limit, whereas the GML estimator is asymptotically efficient yet possesses no optimality property at a finite photon count [19]. Both benchmarks also need to know the PSF, the emitter intensity and the per-pixel noise, and UGIA-F needs the true positions in addition, so UGIA-F is an oracle that a real experiment cannot run whereas EM-GML uses only the system parameters and is practical.

The optimum estimator in the mean square sense is neither of these. It is the Bayes estimator that minimizes the risk over a prior of emitter configurations [22], which is the estimator this paper takes as the target. The Bayes estimator carries a bias and is therefore free of the constraint that makes the CRB diverge, so it stays accurate where the unbiased benchmark does not. For two or more emitters, however, the matched form of the Bayes estimator has no closed form, and its evaluation needs the posterior and hence the full system model, so a direct computation again requires the full system model at every frame. A learned estimator, e.g., a neural network, that approximated the Bayes estimator would therefore be more accurate than the unbiased benchmark at overlap and, once trained, would evaluate the estimate in a single forward pass rather than by the per-frame posterior integration that a direct computation requires.

Neural networks have been applied to SMLM as learned localizers that map a frame directly to emitter positions after training on simulated frames, e.g., the convolutional network Deep-STORM [23], the dense-emitter network DECODE [24], the emission-pattern network smNet [25], the three-dimensional network DeepSTORM3D [26], and the residual deconvolutional network [27]. They remove the per-frame likelihood optimization and run in a single forward pass and they reach high emitter density and high speed. However, their relation to the estimators of estimation theory has not been made explicit, i.e., they are trained and evaluated empirically and are not formulated within the Bayes estimator framework, so it is not established which estimator a frame-trained network approaches, whether the network is optimal in any defined sense, or how it compares with the benchmark estimators of estimation theory. This paper addresses these questions that are theoretically and practically critical when applying neural networks to emitter localization.

The contribution of this paper is twofold. First, we show that a network trained by minimizing an average loss over frames drawn from a prior approaches the Bayes estimator of that loss, and that training with a permutation-matched loss makes the target the matched Bayes estimator whose risk is exactly the localization metric reported here, so the choice of the training loss determines the obtained estimator in the Bayse sense. Second, we test by simulation the hypothesis that such a network outperforms the unbiased oracle UGIA-F and approaches the likelihood estimator EM-GML. Five networks that span learnable activations, Kolmogorov-Arnold edges, convolution, and self attention are trained under a common budget, so that a result common to all of them is attributed to learning from frames rather than to any single design. The simulation confirms the hypothesis on in-distribution frames and locates the advantage of the networks in the overlapped configurations where the oracle UGIA-F diverges.

The networks are trained on frames synthesized with the same system model that EM-GML and UGIA-F use, so the three estimators are built with the same system knowledge, and the networks additionally use the prior over configurations; among the three only UGIA-F uses the true positions and is an oracle, whereas the networks and EM-GML are practically useful, so the accuracy of the networks reflects learning the Bayes-optimal estimator rather than any information advantage. The study is a proof of concept carried out on a field of view (FOV) of $700 \times 700$ nm$^2$. The results and findings motivate and justify the future work to train neural networks on the practically large frames to achieve high-throughput large-FOV super spatiotemporal resolution SMLM.

The paper is organized as follows. Section 2 defines the frame model and the estimators through the Bayes risk framework, i.e., the minimum mean square error (MMSE), the sorted Bayes, the matched Bayes, the GML and the UGIA-F estimators, together with the localization metric. Section 3 defines the neural network estimators and establishes which estimator a frame-trained network approaches. Section 4 computes the matched Bayes estimator by quadrature for two emitters, so that the networks can be measured against the optimum. Section 5 reports the simulation results on an in-distribution dataset and on out-of-distribution circle constellations, together with a stratified analysis by emitter overlap and a verification of the EM-GML benchmark. Section 6 discusses the findings and Section 7 concludes.

## 2. Estimators and the Bayes risk

### *2.1 Data frame model and log-likelihood*

The data frame model of this paper is the universal model of Refs. [15, 17, 19] that is broadly applicable in practical systems and experiments. For simplicity, considered are two-dimensional (2D) imaging, a known emitter number and a known emitter intensity that is the same for all emitters. The analysis and results can be straightforwardly extended to other settings. The model and the estimators are stated here in the notation of Ref. [19] so that the results of the two studies can be read together.

Consider $M$ emitters activated in a data frame. The $m$th emitter is positioned at $\boldsymbol{\theta}_m = (x_m, y_m)^{\mathrm{T}} \in \mathbb{R}^2$ in nm and all positions are stacked in the $2M$ dimensional vector

$$\boldsymbol{\theta} = (\boldsymbol{\theta}_1^{\mathrm{T}}, \dots, \boldsymbol{\theta}_M^{\mathrm{T}})^{\mathrm{T}}. \quad (1)$$

Every emitter emits $I$ photons per second on average and $I$ is known, so that $\beta_m = 1$ for all $m$ in the notation of Ref. [17, 19].

The camera has $K_x \times K_y$ pixels of sizes $\Delta_x$ and $\Delta_y$ in nm, i.e., the FOV is $[0, L_x] \times [0, L_y] = [0, K_x\Delta_x] \times [0, K_y\Delta_y]$, the frame time is $\Delta_t$ in second, the pixel index is $\boldsymbol{k} = (k_x, k_y)$ and the pixel set is $\Omega$. A photon emitted from the $m$th emitter arrives in the camera plane with the PSF, i.e., the probability density funciton $q_m(\boldsymbol{u})$, $\boldsymbol{u} \in \mathbb{R}^2$, whose average over the $\boldsymbol{k}$th pixel is $q_m(\boldsymbol{k})$, $\boldsymbol{k} \in \Omega$. The signal mean, i.e., the mean photon count from all emitters, in the $\boldsymbol{k}$th pixel is

$$s(\boldsymbol{k}) = \Delta_t \Delta_x \Delta_y \, I \, Q(\boldsymbol{k}) \quad (2)$$

where

$$Q(\boldsymbol{k}) = \sum_{m=1}^{M} \beta_m \, q_m(\boldsymbol{k}). \quad (3)$$

The background autofluorescence and the camera readout together produce the mixed noise whose spatiotemporal density in photons per nm$^2$ per second is $b_{\boldsymbol{k}}$ and whose mean photon count in the $\boldsymbol{k}$th pixel is

$$b(\boldsymbol{k}) = \Delta_t \Delta_x \Delta_y \, b_{\boldsymbol{k}}. \quad (4)$$

The mean of the Gaussian readout does not affect the Fisher information and a Gaussian variable whose mean equals its variance is well approximated by a Poisson variable and conversely [15]. The readout mean is therefore set equal to its variance, so that the mixed noise is Poisson and $b_{\boldsymbol{k}}$ denotes the combined background and readout density. Ref. [19] adopts this approximation so that the EM algorithm operates on a Poisson likelihood, and it is adopted here for the same reason. Unlike the earlier studies [15, 17] where $b_{\boldsymbol{k}}$ is spatially uniform, here $b_{\boldsymbol{k}}$ varies from pixel to pixel, which models the slowly spatially varying autofluorescence and the pixel dependent readout of an sCMOS camera [28].

The pixel value is the sum of the signal and the noise,

$$V(\boldsymbol{k}) = S(\boldsymbol{k}) + B(\boldsymbol{k}) \sim \text{Poisson}\big(v(\boldsymbol{k})\big) \tag{5}$$

with

$$v(\boldsymbol{k}) = s(\boldsymbol{k}) + b(\boldsymbol{k}) \tag{6}$$

where $S(\boldsymbol{k})$ and $B(\boldsymbol{k})$ are mutually independent and the pixel values $V(\boldsymbol{k})$ are independent across pixels. The data frame is denoted by $V$ and its probability mass function is

$$f_V(V;\boldsymbol{\theta}) = \prod_{\boldsymbol{k}\in\Omega} \frac{v(\boldsymbol{k})^{V(\boldsymbol{k})} e^{-v(\boldsymbol{k})}}{V(\boldsymbol{k})!}, \tag{7}$$

which is also the likelihood function of $\boldsymbol{\theta}$. Dropping the constant that is independent of $\boldsymbol{\theta}$, the per-frame log-likelihood is

$$\ell_1(\boldsymbol{\theta}) = \sum_{\boldsymbol{k}\in\Omega} [V(\boldsymbol{k}) \ln v\,(\boldsymbol{k}) - v(\boldsymbol{k})] \tag{8}$$

and the Fisher information matrix [15, 19] is

$$\mathbf{F}(\boldsymbol{\theta}) = \sum_{\boldsymbol{k}\in\Omega} \frac{1}{v(\boldsymbol{k})} \frac{\partial v(\boldsymbol{k})}{\partial \boldsymbol{\theta}} \frac{\partial v(\boldsymbol{k})}{\partial \boldsymbol{\theta}^{\mathrm{T}}}. \tag{9}$$

An estimator estimates $\boldsymbol{\theta}$ from the single frame $V$, that is, an estimator is a function $\hat{\boldsymbol{\theta}}(V)$.

### 2.2 Bayes risk

The estimators compared in this paper are constructed on different principles, so a common criterion is needed before any of them can be judged. The criterion adopted here is the Bayes risk, which is the average estimation error over both the noise in the frame and the emitter positions that produced it.

Let $p(\boldsymbol{\theta})$ denote the prior, that is, the distribution from which the emitter positions are drawn. When $\boldsymbol{\theta}$ is regarded as random with prior $p(\boldsymbol{\theta})$, the function $f_V(V;\boldsymbol{\theta})$ of Eq. (7) is the conditional probability mass function of $V$ given $\boldsymbol{\theta}$, so the joint distribution of the frame and the emitter positions is $f_V(V;\boldsymbol{\theta})p(\boldsymbol{\theta})$. Let $L\big(\hat{\boldsymbol{\theta}},\boldsymbol{\theta}\big)$ denote a loss function that measures the error of an estimate. The risk of an estimator is its loss averaged over the joint distribution,

$$R\big(\hat{\boldsymbol{\theta}}\big) = \int\int L\left(\hat{\boldsymbol{\theta}}(V),\boldsymbol{\theta}\right) f_V(V;\boldsymbol{\theta})\, p(\boldsymbol{\theta})\, dV\, d\boldsymbol{\theta}. \tag{10}$$

The Bayes estimator is defined as the estimator that minimizes $R$ over all functions $\hat{\boldsymbol{\theta}}(\cdot)$ for the loss $L$ and the prior $p$. Its risk is the smallest attainable by any estimator whatsoever, so it is the bottom line against which every other estimator can be measured.

The Bayes estimator has a simple characterization. Let

$$p(V) = \int f_V\,(V;\boldsymbol{\theta})\, p(\boldsymbol{\theta})\, d\boldsymbol{\theta} \tag{11}$$

be the marginal distribution of the frame, that is, the distribution of a frame produced by an emitter configuration drawn from the prior. The posterior of the emitter positions given the observed frame is then

$$p(\boldsymbol{\theta} \mid V) = \frac{f_V(V;\boldsymbol{\theta})\, p(\boldsymbol{\theta})}{p(V)}, \tag{12}$$

so that the joint distribution factors as

$$f_V(V;\boldsymbol{\theta})\, p(\boldsymbol{\theta}) = p(\boldsymbol{\theta} \mid V)\, p(V). \tag{13}$$

Substituting Eq. (13) into Eq. (10) and integrating over $\boldsymbol{\theta}$ first gives

$$R(\hat{\boldsymbol{\theta}}) = \int \left[\int L\left(\hat{\boldsymbol{\theta}}(V), \boldsymbol{\theta}\right) p(\boldsymbol{\theta} \mid V)\, d\boldsymbol{\theta}\right] p(V)\, dV. \tag{14}$$

The inner integral of Eq. (14) depends on the estimator only through the value $\hat{\boldsymbol{\theta}}(V)$ that the estimator takes at the observed frame, and $p(V)$ is non-negative, so the risk is minimized by minimizing the inner integral separately for every $V$. The Bayes estimator is therefore

$$\hat{\boldsymbol{\theta}}_{\text{Bayes}}(V) = \arg\min_{\boldsymbol{u}} \int L\,(\boldsymbol{u}, \boldsymbol{\theta})\ p(\boldsymbol{\theta} \mid V)\, d\boldsymbol{\theta}, \tag{15}$$

which is computed frame by frame although it minimizes an average over frames.

Two consequences are used throughout this paper. First, the Bayes estimator depends on the loss function, so different losses define different optimal estimators even for the same data. Second, the optimality holds only when the prior in the posterior is the distribution the frames were actually drawn from, so the guarantee is lost when an estimator built for one prior is applied to data generated by another.

Sections 2.3 to 2.5 derive the Bayes estimator for three loss functions. Each of the three is a loss with which a neural network can be trained, so the three subsections together state which estimator a network approaches for a given choice of the training loss, and thereby the principle by which a network is to be trained. Sections 2.6 and 2.7 then present the GML and UGIA-F estimators. Neither of the two is a Bayes estimator, and both are used in this paper only as benchmarks against which the networks are compared.

### *2.3 Minimum mean square error estimator*

The natural loss for localization is the squared error

$$L_{\text{SE}}(\boldsymbol{u}, \boldsymbol{\theta}) = \sum_{m=1}^{M} \|\boldsymbol{u}_m - \boldsymbol{\theta}_m\|^2\,, \tag{16}$$

so the Bayes estimator of Eq. (15) minimizes the mean square error (MSE), that is $\mathbb{E}[L_{\text{SE}}(\boldsymbol{u}, \boldsymbol{\theta}) \mid V] = \mathbb{E}[\|\boldsymbol{u} - \boldsymbol{\theta}\|^2 \mid V]$, where the expectation is taken with respect to $\boldsymbol{\theta}$ conditioned on $V$. It is easy to obtain that the minimum mean square error (MMSE) estimator is the conditional mean

$$\hat{\boldsymbol{\theta}}_{\text{MMSE}}(V) = \mathbb{E}[\boldsymbol{\theta} \mid V]. \tag{17}$$

This estimator is degenerate for multiple emitters, as the following proposition shows.

*Proposition 1:* Let $\mathcal{S}_M$ be the set of permutations of $\{1, \dots, M\}$ and for $\pi \in \mathcal{S}_M$ let $\mathbf{P}_\pi$ permute the emitter blocks of $\boldsymbol{\theta}$. If the prior is exchangeable, that is $p(\mathbf{P}_\pi \boldsymbol{\theta}) = p(\boldsymbol{\theta})$ for all $\pi \in \mathcal{S}_M$, then

$$\mathbb{E}[\boldsymbol{\theta}_1 \mid V] = \mathbb{E}[\boldsymbol{\theta}_2 \mid V] = \cdots = \mathbb{E}[\boldsymbol{\theta}_M \mid V]. \tag{18}$$

is yielded by the MMSE estimator. ■

The proof is given in Appendix A.1.

Since the emitters in an experiment are physically indistinguishable, the prior is exchangeable whenever the emitter positions are drawn independently, so Proposition 1 applies to the setting of this paper, i.e., emitter positions drawn independently, and to any other setting in which the emitters are not labeled by some further observable. All $M$ components of $\hat{\boldsymbol{\theta}}_{\mathrm{MMSE}}$ coincide at the common posterior mean of Eq. (18), so the MMSE estimator reports a single point regardless of how the emitters are arranged and it can never resolve them.

The reason for the degeneracy is the mismatch between what the loss requires and what the frame supplies. The loss $L_{\mathrm{SE}}$ of Eq. (16) is defined with respect to the emitter indices, since the $m$th estimate is penalized by its distance to the $m$th emitter and to no other. The frame however carries no information that distinguishes one index from another, because the likelihood of Eq. (7) is invariant under any permutation of the indices. The posterior marginal of $\boldsymbol{\theta}_m$ is therefore the same for every $m$, and the value of $\boldsymbol{u}_m$ that minimizes $\mathbb{E}[\|\boldsymbol{u}_m - \boldsymbol{\theta}_m\|^2 \mid V]$ is that one common posterior mean for every $m$. In short, the loss demands a correspondence between estimates and emitters that the posterior does not provide, and the estimator answers by giving the same estimate to all of them.

For $M = 2$ the resulting MMSE has a closed form.

*Corollary 1:* Let $M = 2$. Denote by $d = \|\boldsymbol{\theta}_1 - \boldsymbol{\theta}_2\|$ the emitter separation, $\boldsymbol{c} = (\boldsymbol{\theta}_1 + \boldsymbol{\theta}_2)/2$ the midpoint of the two emitters and $\hat{\boldsymbol{m}}$ the common posterior mean of Eq. (18). Then

$$\sum_{m=1}^{2} \|\hat{\boldsymbol{m}} - \boldsymbol{\theta}_m\|^2 = \frac{d^2}{2} + 2\|\hat{\boldsymbol{m}} - \boldsymbol{c}\|^2 \tag{19}$$

is the MMSE of Eq. (16) ■

The proof is given in Appendix A.2.

The degenerate estimator is therefore the centroid estimator, whose error is set by half the emitter separation and is small only when the emitters are unresolvable.

### *2.4 The sorted Bayes estimator*

The degeneracy of Section 2.3 follows from the exchangeability of the posterior, so it can be removed by breaking that exchangeability. The simplest way is to impose a canonical order on the emitters, for instance by their $x$ coordinates. Let $\boldsymbol{\theta}_{(1)}, \dots, \boldsymbol{\theta}_{(M)}$ denote the emitter positions arranged so that $x_{(1)} \le \cdots \le x_{(M)}$ and consider the sorted squared error

$$L_{\mathrm{S}}(\boldsymbol{u}, \boldsymbol{\theta}) = \sum_{m=1}^{M} \left\|\boldsymbol{u}_m - \boldsymbol{\theta}_{(m)}\right\|^2. \tag{20}$$

The prior of the sorted vector is not exchangeable, so Proposition 1 no longer applies. It is easy to obtain that the Bayes estimator of Eq. (15) for this loss is

$$\hat{\boldsymbol{\theta}}_{\mathrm{S}}(V) = \mathbb{E}\left[\boldsymbol{\theta}_{(\cdot)} \mid V\right], \tag{21}$$

returning $M$ distinct estimates. It is the posterior mean of the order statistics of the emitter positions along the $x$ axis.

The sorting removes the degeneracy at a price, which is an anisotropy between the sorting coordinate $x$ and the coordinate $y$ orthogonal to it. The order is decided by the $x$ coordinates alone, so whenever two emitters have close $x$ coordinates, the noise decides which of them is called the first, and the $y$ estimate that accompanies the decision jumps between the two emitters. The following proposition quantifies the effect for two emitters.

*Proposition 2:* Let $M = 2$ and let the two estimates be ordered by their $x$ coordinates. For a fixed true configuration let the unordered estimates over repeated frames be $X_m = x_m + \xi_m$ and $Y_m = y_m + \eta_m$ for $m = 1, 2$, where $\xi_1, \xi_2, \eta_1, \eta_2$ are mutually independent and $\mathcal{N}(0, \sigma^2)$. Write $d_x = |x_1 - x_2|$ and $d_y = |y_1 - y_2|$ for the coordinate separations of the two emitters and thus

$$p = \Phi\left(-\frac{d_x}{\sigma\sqrt{2}}\right) \tag{22}$$

is the probability that the order of $X_1$ and $X_2$ disagrees with the order of $x_1$ and $x_2$, where $\Phi$ is the standard normal distribution function. Then the ordered estimate of the $y$ coordinate has variance

$$\mathrm{Var}\big(Y_{(1)}\big) = \sigma^2 + p(1-p)\, d_y^2 \tag{23}$$

whereas the ordered estimate of the $x$ coordinate has variance determined by the mean absolute difference of the two estimates,

$$\mathbb{E}\,[|X_1 - X_2|] = \frac{2\sigma}{\sqrt{\pi}} \exp\left(-\frac{d_x^2}{4\sigma^2}\right) + d_x\,(1 - 2p), \tag{24}$$

namely

$$\mathrm{Var}\big(X_{(1)}\big) = \sigma^2 + \frac{1}{4}[d_x^2 - \mathbb{E}^2\,(|X_1 - X_2|)], \tag{25}$$

which increases monotonically from $\sigma^2(1 - 1/\pi)$ at $d_x = 0$ to $\sigma^2$ as $d_x \to \infty$. ■

The proof is given in Appendix A.3.

Two consequences follow. The excess variance $p(1-p)d_y^2$ appears only in the coordinate that is not used for sorting, so the estimates of one emitter obtained from repeated frames scatter more widely along $y$ than along $x$. And the excess is largest when the two emitters are aligned perpendicular to the sorting axis, i.e., $d_x \ll d_y$. Since $d_x = 0$ gives $p = 1/2$ and

$$\mathrm{Var}\big(Y_{(1)}\big) = \sigma^2 + \frac{d_y^2}{4}, \tag{26}$$

the $y$ error, i.e., the square root of the variance of Eq. (26), approaches $d_y/2$, which is the error of the degenerate MMSE estimator in Corollary 1. In that geometry sorting therefore recovers nothing beyond the degenerate collapse that it was introduced to avoid. Conversely, when $d_x$ is large compared with $\sigma$, the probability $p$ vanishes and the sorted estimator agrees with the matched estimator of Section 2.5, implying the sorting on the $x$ coordinates alone is sufficient. For $M > 2$ the same anisotropy arises for every pair of emitters whose $x$ coordinates are close. Therefore, sorting on the $x$ coordinates alone still causes ambiguity in the estimates of $y$ coordinates particularly when the $x$ coordinates are close.

### 2.5 The matched Bayes estimator

The sorted loss of Eq. (20) breaks the exchangeability by an arbitrary convention rather than by the geometry of the estimate, which is the origin of the anisotropy of Proposition 2. The alternative is to let the correspondence between estimates and emitters be chosen so as to minimize the error itself, which gives the permutation matched squared error

$$L_{\mathrm{M}}(\boldsymbol{u}, \boldsymbol{\theta}) = \min_{\pi \in \mathcal{S}_M} \sum_{m=1}^{M} \left\|\boldsymbol{u}_m - \boldsymbol{\theta}_{\pi(m)}\right\|^2, \tag{27}$$

whose Bayes estimator is

$$\hat{\boldsymbol{\theta}}_{\mathrm{B}}(V)=\arg\min_{\boldsymbol{u}}\int\left[\min_{\pi\in\mathcal{S}_M}\sum_{m=1}^{M}\left\|\boldsymbol{u}_m-\boldsymbol{\theta}_{\pi(m)}\right\|^2\right]p(\boldsymbol{\theta}\mid V)\,d\boldsymbol{\theta}. \tag{28}$$

Here $\pi(m)$ is the index of the true emitter matched to the $m$th estimate by the permutation $\pi\in\mathcal{S}_M$.

The minimization over $\pi$ inside the integral removes the dependence of the loss on the emitter indices, so the exchangeability of the posterior no longer forces the estimates together, and $\hat{\boldsymbol{\theta}}_{\mathrm{B}}$ returns $M$ distinct estimates. The minimizing permutation is the Hungarian assignment [29]. This is the reason why the Hungarian assignment is required in the training of a network. Without it the training loss is $L_{\mathrm{SE}}$ of Eq. (16), whose Bayes estimator is the degenerate single posterior mean of Proposition 1, so a network trained without the assignment converges to $M$ coincident estimates and can never resolve the emitters. The sorting on the $x$ coordinates alone also yields large variances on the estimates of $y$ coordinates as indicated in Proposition 2. The Hungarian assignment is the sorting on both $x$ and $y$ coordinates, thus completely eliminating the index ambiguity. For $M=1$ the assignment is trivial and $\hat{\boldsymbol{\theta}}_{\mathrm{B}}$ reduces to the posterior mean.

$\hat{\boldsymbol{\theta}}_{\mathrm{B}}$ has no closed form for $M\geq 2$ and is computed in Section 4 by quadrature over the posterior for $M=2$. It is the central object of this paper, since it is the minimizer of the risk under the loss that the quality metric of Section 2.8 measures, so no estimator can achieve a smaller value of that metric on frames drawn from the prior. In particular, it is not worse than either of the two estimators of Sections 2.3 and 2.4.

The absence of a closed form is due to the minimization over permutations in Eq. (27), i.e., the winning permutation depends on the estimate so the objective of Eq. (28) is a non-convex piecewise quadratic function of the estimate whose minimizer cannot be written in closed form for $M\geq 2$. The estimate is instead obtained by the fixed-point iteration of Section 4, i.e., an assignment step that matches the posterior mass to the current estimate and an update step that sets each estimate to the posterior mean of the mass assigned to it, alternated to convergence. The iteration also needs the posterior $p(\boldsymbol{\theta}\mid V)$ of Eq. (12) and hence the likelihood of Eq. (7) with its PSF, its intensity and its per-pixel noise, so a direct computation of $\hat{\boldsymbol{\theta}}_{\mathrm{B}}$ uses the full system model at every frame. This per-frame computational cost is one motivation for approximating $\hat{\boldsymbol{\theta}}_{\mathrm{B}}$ by a network that, once trained, evaluates the estimate in a single forward pass, which is developed in Section 3.

*Corollary 2:* Let $\hat{\boldsymbol{\theta}}_{\mathrm{MMSE}}$, $\hat{\boldsymbol{\theta}}_{\mathrm{S}}$ and $\hat{\boldsymbol{\theta}}_{\mathrm{B}}$ be the estimators of Eqs. (17), (21) and (28). Then

$$R(\hat{\boldsymbol{\theta}}_{\mathrm{B}})\leq R(\hat{\boldsymbol{\theta}}_{\mathrm{MMSE}}),\quad R(\hat{\boldsymbol{\theta}}_{\mathrm{B}})\leq R(\hat{\boldsymbol{\theta}}_{\mathrm{S}}) \tag{29}$$

under the matched loss $L_{\mathrm{M}}$ of Eq. (27). ■

The proof is given in Appendix A.4.

The inequalities of Eq. (29) are not strict. By Proposition 2 the sorted estimator agrees with the matched estimator when the $x$ coordinates of every pair of emitters are well separated compared with $\sigma$, so the two risks are equal in that limit and the gap opens only as emitters approach one another in the sorting coordinate. The corollary compares the risks, that is the averages over the prior, and it does not assert that $\hat{\boldsymbol{\theta}}_{\mathrm{B}}$ is more accurate than the other two on every individual frame.

### *2.6 Global maximum likelihood estimator*

The GML estimator [19] maximizes the per-frame log-likelihood,

$$\hat{\boldsymbol{\theta}}_{\mathrm{GML}}=\arg\max_{\boldsymbol{\theta}}\ell_1(\boldsymbol{\theta}). \tag{30}$$

It is asymptotically consistent, asymptotically normal and asymptotically efficient as the Fisher information grows without bound [19], whereas it possesses no optimality property at a finite photon count. In the Bayes framework of Section 2.2, $\hat{\boldsymbol{\theta}}_{\mathrm{GML}}$ is the mode of the posterior under a flat prior, that is under $p(\boldsymbol{\theta}) = \mathrm{const}$, which is a different functional of the posterior from the conditional mean of Eq. (17).

The expectation maximization (EM) algorithm [19] performs a local search that increases $\ell_1(\boldsymbol{\theta})$ at every iteration and converges to a local maximum likelihood (LML) point. The likelihood function of multiple emitters may present several LML points, one of which is the global maximizer $\hat{\boldsymbol{\theta}}_{\mathrm{GML}}$, so a single EM run started at an arbitrary point converges to an LML point rather than to $\hat{\boldsymbol{\theta}}_{\mathrm{GML}}$. When the emitters overlap severely, the Fisher information matrix is near singular, so the log-likelihood is nearly flat about its maximizer and the ascent is slow, which further reduces the chance that a single run arrives at the global maximizer within a finite iteration budget.

The global maximizer is therefore approached by multi-start EM. Let $\hat{\boldsymbol{\theta}}^{(1)}, \dots, \hat{\boldsymbol{\theta}}^{(K)}$ be the LML points to which $K$ EM runs converge from $K$ independent random initial positions and let

$$k^{\star} = \arg \max_{1 \le k \le K} \ell_1\left(\hat{\boldsymbol{\theta}}^{(k)}\right). \tag{31}$$

The EM-GML estimate is the one that achieves the largest log-likelihood among the $K$ points,

$$\hat{\boldsymbol{\theta}}_{\text{EM-GML}} = \hat{\boldsymbol{\theta}}^{(k^{\star})}, \tag{32}$$

which is taken as an approximation of $\hat{\boldsymbol{\theta}}_{\mathrm{GML}}$.

Selection by Eq. (31) uses the log-likelihood alone, which depends only on the estimated positions and the observed frame, so the selection rule requires no knowledge of the true positions and a larger $K$ can only bring the estimate closer to $\hat{\boldsymbol{\theta}}_{\mathrm{GML}}$. The initial positions are drawn in a neighborhood of the true positions so that the multi-start search reliably reaches the global maximizer for the benchmark, whereas the global maximizer itself is a function of the frame and the system parameters and does not use the true positions, so EM-GML is a practical estimator. Section 5.6 verifies against an exhaustive search over the whole parameter space that $K = 250$ attains the global maximizer on 99.7 percent of frames for $M = 2$. In practice the true positions are unavailable and EM can be initialized by a successive interference cancellation (SIC) estimator [21] or any other estimator at the cost of losing the guarantee.

### *2.7 Unbiased Gaussian information achieving estimator*

The Cramér-Rao bound (CRB) is the matrix $\mathbf{F}(\boldsymbol{\theta})^{-1}$ and the covariance of any unbiased estimator is bounded below by it in the Loewner ordering. The UGIA-F estimator [18, 19] is the Gaussian estimator

$$\hat{\boldsymbol{\theta}}_{\mathrm{F}} \sim \mathcal{N}(\boldsymbol{\theta}, \mathbf{F}(\boldsymbol{\theta})^{-1}), \tag{33}$$

which is unbiased and attains the Fisher information and the CRB. It is realized by

$$\hat{\boldsymbol{\theta}}_{\mathrm{F}} = \boldsymbol{\theta} + \mathbf{F}(\boldsymbol{\theta})^{-1/2} \boldsymbol{g} \tag{34}$$

where $\mathbf{F}(\boldsymbol{\theta})^{-1/2}$ is the inverse square root of the Fisher information matrix and $\boldsymbol{g}$ is a standard Gaussian vector. The UGIA-F estimator is an oracle benchmark rather than a deployable algorithm since its construction uses the true positions, and it represents the accuracy attainable by the best unbiased estimator.

Unbiasedness is a constraint, not an optimality property. When two emitters approach each other, the Fisher information matrix becomes near singular and the CRB diverges, so the accuracy attainable under the unbiasedness constraint degrades without limit while a biased estimator such as the above EM-GML [19] remains free of that constraint.

### *2.8 Quality metric*

Every estimator of Sections 2.3 to 2.7 produces $M$ estimated positions from one data frame, whereas the indices of the estimates are unknown, so the root mean square error (RMSE), defined as $\left[L_{\mathrm{SE}}(\hat{\boldsymbol{\theta}}, \boldsymbol{\theta})/M\right]^{1/2}$ where $L_{\mathrm{SE}}$ is the squared error of Eq. (16) evaluated under the true correspondence, cannot be evaluated and a correspondence between the estimated and the true positions must be established before an error can be computed.

The correspondence is chosen by maximum likelihood. Under the model in which every estimate is Gaussian about its true position with a common standard deviation $\sigma$, which holds approximately for any reasonable estimator [30], the likelihood of the correspondence $\pi$ is

$$\ln L\,(\pi) = \mathrm{const} - \frac{1}{2\sigma^2} \sum_{m=1}^{M} \left\| \hat{\boldsymbol{\theta}}_{\pi(m)} - \boldsymbol{\theta}_m \right\|^2, \tag{35}$$

so the correspondence of maximum likelihood is the one that minimizes the total squared distance, which is the assignment computed by the Hungarian algorithm. The quality metric is therefore the root mean square minimum distance under that assignment, denoted RMSMD-H,

$$D_{\mathrm{H}}(\hat{\boldsymbol{\theta}}, \boldsymbol{\theta}) = \left[ \frac{1}{M} \min_{\pi \in \mathcal{S}_M} \sum_{m=1}^{M} \left\| \hat{\boldsymbol{\theta}}_{\pi(m)} - \boldsymbol{\theta}_m \right\|^2 \right]^{1/2}. \tag{36}$$

Two properties of Eq. (36) are used repeatedly. First, $MD_{\mathrm{H}}^2$ is exactly the loss $L_{\mathrm{M}}$ of Eq. (27), so the matched Bayes estimator $\hat{\boldsymbol{\theta}}_{\mathrm{B}}$ of Eq. (28) is by construction the minimizer of the reported metric Eq. (36), and the training loss of Section 3.3 evaluates the same quantity. Second, since the minimization is over assignments, $D_{\mathrm{H}}$ is not greater than the RMSE evaluated under the true correspondence where $\pi$ is an identity permutation.

A data movie of $N$ frames is processed frame by frame, and the metric reported for the movie is the average RMSMD-H, denoted ARMSMD-H,

$$\overline{D}_{\mathrm{H}} = \left[ \frac{1}{N} \sum_{n=1}^{N} D_{\mathrm{H}}^2\left(\hat{\boldsymbol{\theta}}_n, \boldsymbol{\theta}_n\right) \right]^{1/2}, \tag{37}$$

which is the Monte Carlo estimate of $R(\hat{\boldsymbol{\theta}})/M$ under the loss $L_{\mathrm{M}}$. The averaging is performed on the squared distances and the square root is taken once at the end.

The accuracy of an estimator reported in this paper is its RMSMD-H and ARMSMD-H of Eqs. (36) and (37), i.e., the smaller the value the more accurate. RMSMD-H and ARMSMD-H measure the average distance between the estimate and its true position on a 2D plane. When considering the average error in one axis as usually considered in microscopy, the metrics shall be divided by $\sqrt{2}$.

The metric is defined frame by frame because this paper estimates the emitter positions from a single data frame, so every frame is an independent estimation problem in which the number of estimates equals the number of emitters. The root mean square minimum distance RMSMD [18] and its partition form RMSMD-P [30] evaluate a reconstructed SMLM image instead, in which the estimates of all frames of a movie are pooled and one emitter carries many estimates, so the number of estimates exceeds the number of emitters and no correspondence between the two sets exists. They are not used here. The companion study [19] reports the RMSE because there the indices of the estimates of both the UGIA-F estimator and the EM-GML estimator are known, whereas here the neural networks produce unindexed estimates so the RMSE is unavailable and $\overline{D}_{\mathrm{H}}$ is used for all estimators so that they are compared on the same footing.

## 3. Neural network estimators

### *3.1 What a frame-trained network estimates*

A neural network estimator is a function $\hat{\boldsymbol{\theta}}_{\mathrm{NN}}(V;\boldsymbol{w})$ of the frame, parameterized by the network weights $\boldsymbol{w}$. The weights are chosen to minimize the empirical average of a loss $L$ over a set of training frames,

$$\hat{\boldsymbol{w}} = \arg\min_{\boldsymbol{w}} \frac{1}{N_{\mathrm{t}}} \sum_{n=1}^{N_{\mathrm{t}}} L\left(\hat{\boldsymbol{\theta}}_{\mathrm{NN}}(V_n;\boldsymbol{w}), \boldsymbol{\theta}_n\right), \tag{38}$$

where the $N_{\mathrm{t}}$ training pairs $(\boldsymbol{\theta}_n, V_n)$ are generated by drawing $\boldsymbol{\theta}_n$ from the prior $p(\boldsymbol{\theta})$ and then $V_n$ from the frame model of Eq. (5).

Eq. (38) is a sample version of the risk $R$ of Eq. (10) under the loss $L$. As $N_{\mathrm{t}}$ grows, the sample average converges to $R$, and the minimizer of $R$ over all functions of the frame is the Bayes estimator for that loss by Eq. (15). A frame-trained network therefore approximates the Bayes estimator of whichever loss it is trained with, and it departs from the Bayes estimator only through the finite capacity of the network, the finite number of training frames and the optimization. This is the theoretical basis on which a neural network is used as an estimator in this paper, and it is what distinguishes a network from the GML and UGIA-F estimators, which target the posterior mode and the best unbiased accuracy respectively.

The three losses of Sections 2.3 to 2.5 therefore give three different networks from the same architecture and the same training frames. First, a network trained with the squared error $L_{\mathrm{SE}}$ of Eq. (16) approaches the MMSE estimator of Eq. (17), which by Proposition 1 is degenerate, so its $M$ outputs converge to a single point and the emitters are never resolved. This holds whenever the emitter labels supplied in training are exchangeable, which is the case when the emitter positions are drawn independently and presented in the order generated. Second, a network trained with the sorted squared error $L_{\mathrm{S}}$ of Eq. (20) approaches the sorted Bayes estimator of Eq. (21). Its outputs are distinct, whereas by Proposition 2 its estimates scatter more widely along the coordinate that is not used for sorting, and the excess is largest for emitter pairs aligned perpendicular to the sorting axis. Third, a network trained with the matched squared error $L_{\mathrm{M}}$ of Eq. (27) approaches the matched Bayes estimator of Eq. (28). Its outputs are distinct and, by Corollary 2, its risk under the reported metric is not greater than that of either of the other two. The choice of the training loss therefore determines the estimator that is obtained, and the matched loss is chosen in this paper because it is the loss whose Bayes estimator minimizes exactly the metric of Section 2.8. The three networks are compared numerically in Section 5.

Three further properties of Eq. (38) are worth stating explicitly. First, the network is built with the system parameters through its training frames. The training pairs of Eq. (38) are generated from the frame model of Eq. (5) with the PSF, the emitter intensity and the per-pixel noise, and from the true positions drawn from the prior, so the network is given the same system knowledge as the GML and UGIA-F estimators, i.e., it acquires that knowledge in training rather than by an explicit per-frame computation. Once trained, it maps a frame to positions in a single forward pass and does not recompute the likelihood. Second, the network additionally uses the prior. Training draws $\boldsymbol{\theta}_n$ from $p(\boldsymbol{\theta})$, so the network learns the distribution of emitter configurations, which the GML and UGIA-F estimators do not use. The network therefore has the same system knowledge as EM-GML and UGIA-F and the prior in addition, which is the side information that the Bayes framework prescribes and is examined in Section 6. Third, the prior used in training is uniform. When the distribution of the emitter positions is unknown, the least informative choice on a bounded support $\mathcal{B}$ is the uniform density, i.e., $p(\boldsymbol{\theta})$ constant on $\mathcal{B}$ and zero outside, which is the maximum entropy prior on $\mathcal{B}$ and commits to no particular arrangement of the emitters. Training on the uniform prior therefore makes the network accurate across the whole support rather than tuned to an assumed clustering, whereas a more informative prior can replace it whenever one is known.

Under the uniform prior, the estimator reduces to a likelihood weighted average. With $p(\boldsymbol{\theta})$ constant on $\mathcal{B}$, the posterior of Eq. (12) becomes the likelihood of Eq. (7) normalized over $\mathcal{B}$, i.e., for $\boldsymbol{\theta} \in \mathcal{B}$,

$$p(\boldsymbol{\theta} \mid V) = \frac{f_V(V; \boldsymbol{\theta})}{\int_{\mathcal{B}} f_V\,(V; \boldsymbol{\theta}')\, d\boldsymbol{\theta}'}, \tag{39}$$

and the posterior is zero outside $\mathcal{B}$. The matched Bayes estimator of Eq. (28) is therefore

$$\hat{\boldsymbol{\theta}}_{\mathrm{B}}(V) = \arg\min_{\boldsymbol{u}} \int_{\mathcal{B}} L_{\mathrm{M}}\,(\boldsymbol{u}, \boldsymbol{\theta})\; f_V(V; \boldsymbol{\theta})\, d\boldsymbol{\theta}, \tag{40}$$

i.e., the matched average of the emitter positions under the likelihood restricted to $\mathcal{B}$. This estimator is the matched mean of the normalized likelihood whereas the global maximum likelihood estimator of Eq. (30) is its mode, i.e., the two are different functionals of the same normalized likelihood. A network trained on frames drawn from the uniform prior therefore approaches the matched mean of the normalized likelihood whereas EM-GML reaches its mode, so the two differ as the mean and the mode of the same normalized likelihood, which is the origin of the difference between them observed in Section 5.

The network approximation of $\hat{\boldsymbol{\theta}}_{\mathrm{B}}$ is valid only on the prior it was trained on. By the second consequence stated in Section 2.2, the optimality of $\hat{\boldsymbol{\theta}}_{\mathrm{B}}$ and hence the accuracy of the network hold for frames drawn from $p(\boldsymbol{\theta})$, so all comparisons in this paper are made on such frames.

### *3.2 Network architectures*

Five networks are compared. They are not proposed as new architectures for localization and no claim is made that any of them is specific to or optimal for the problem. They are chosen to span a range of inductive biases, so that a result common to all of them can be attributed to learning from frames rather than to the design of any one network. They form four levels.

**Level 1, learnable activations on a plain network.** The adaptive piecewise linear network, denoted APL, applies a linear map followed by a learnable piecewise linear activation built on a grid of knots [31, 32]. It is the oldest and simplest of the five and is included as a deliberate plain baseline, that is, as the network least likely to possess any advantage for this problem.

**Level 2, learnable edge functions.** The Kolmogorov-Arnold network, denoted KAN, replaces the linear map by a learnable univariate function on every edge, each function being the sum of a sigmoid linear unit residual and a B-spline on a fixed grid [33]. KAN is a popularly applied novel network architecture. It is a different and more expressive functional basis than APL at a comparable depth.

**Level 3, convolutional front end.** The networks CNN-APL and CNN-KAN, where CNN stands for convolutional neural network [34], prepend a two layer convolutional stem to the APL and KAN heads respectively. The stem supplies spatial translation aware features that neither plain network can form directly, thus improving training speed and localization accuracy.

**Level 4, self attention.** The vision transformer, denoted ViT [35], treats each pixel of the frame as one token, prepends a learnable class token and applies a stack of transformer encoder blocks with multi-head self attention. The class token output is mapped to the $M$ estimated positions. Self attention captures spatial correlations of a frame in a manner analogous to convolution, whereas it uses no fixed kernel and no built-in translation equivariance.

All five networks accept the frame as a vector of $K_x K_y$ pixel values normalized by the largest value of the frame and produce $2M$ numbers that are read as $M$ positions in nanometers. The normalization removes the absolute photon scale from the input, so the networks are given the shape of the frame rather than its photon count.

The reader is referred to the references for the detailed architectures of the five networks.

### 3.3 Training

All five networks are trained by minimizing Eq. (38) with the matched loss $L_{\mathrm{M}}$ of Eq. (27) evaluated by the Hungarian assignment. The assignment is recomputed for every frame at every step, so no ordering of the estimates is imposed and the network is free to produce the $M$ positions in any order.

Training frames are generated on the fly, that is, a fresh batch of emitter configurations is drawn from the prior and a fresh batch of frames is generated at every step, so no training frame is presented twice and $N_{\mathrm{t}}$ in Eq. (38) equals the total number of frames generated. The optimizer is Adam [36] with a cosine annealing learning rate [37]. All networks are trained with the same batch size, the same number of steps and therefore the same number of training frames, so that their accuracies and their convergence can be compared on an equal budget. A separate network is trained for each emitter number $M$.

## 4. Performance of Bayes estimators for two emitters

The matched Bayes estimator $\hat{\boldsymbol{\theta}}_{\mathrm{B}}$ of Eq. (28) has no closed form for $M \geq 2$ and is the minimizer of the reported metric, so it must be computed before any estimator can be measured against it. We consider the case of $M = 2$ where the complexity of numerical computations is acceptable.

For $M = 2$ the space of emitter positions $\boldsymbol{\theta}$ is four dimensional and the posterior can be integrated numerically by quadrature, which yields $\hat{\boldsymbol{\theta}}_{\mathrm{B}}$ and the degenerate MMSE estimator $\hat{\boldsymbol{\theta}}_{\mathrm{MMSE}}$ of Eq. (17) exactly up to the grid in the numerical evaluation. This section states the computation, whereas the validation that no estimator falls below $\hat{\boldsymbol{\theta}}_{\mathrm{B}}$ on frames drawn from the prior is reported in Section 5.

### 4.1 The posterior on a grid

The prior $p(\boldsymbol{\theta})$ is uniform on the central $3 \times 3$ pixel block $\mathcal{B}$ of FOV, both emitters are drawn independently and uniformly from that region, so by Eq. (12) the posterior is proportional to the likelihood and no prior factor enters the computation. The region $\mathcal{B}$ is covered by a square grid of $S = G^2$ single emitter nodes $\boldsymbol{\xi}_1, \ldots, \boldsymbol{\xi}_S$ of spacing $\Delta_g$ where $G = 3\Delta_x/\Delta_g$ with $\Delta_x = \Delta_y$ is the number of nodes in one axis.

The posterior being discretized is the joint over the two labeled emitter positions $(\boldsymbol{\theta}_1, \boldsymbol{\theta}_2)$, whose support is the product $\mathcal{B} \times \mathcal{B}$, i.e. emitter 1 has its own coordinate and emitter 2 has its own. The two emitters are placed at ordered pairs of nodes. A grid point of that product space assigns emitter 1 to node $\boldsymbol{\xi}_a$ and emitter 2 to node $\boldsymbol{\xi}_b$, so $(\boldsymbol{\xi}_a, \boldsymbol{\xi}_b)$ and $(\boldsymbol{\xi}_b, \boldsymbol{\xi}_a)$ are two distinct points with two different label assignments. There are exactly $S^2$ such ordered pairs. For example, for a $3 \times 3$ pixel block $\mathcal{B}$, if $\Delta_x = \Delta_y = 100$ nm, $\Delta_g = 6$ nm, then $G = 50$ and there are $S = 2{,}500$ nodes for one emitter and $S^2 = 6{,}250{,}000$ ordered pairs of nodes for two emitters.

For an emitter at node $\boldsymbol{\xi}$, the signal mean in the $\boldsymbol{k}$th pixel is $s(\boldsymbol{k}; \boldsymbol{\xi})$ of Eq. (2), so by Eqs. (3) and (6) the mean photon count of the $\boldsymbol{k}$th pixel by the pair $(\boldsymbol{\xi}_a, \boldsymbol{\xi}_b)$ is

$$v(\boldsymbol{k}; \boldsymbol{\xi}_a, \boldsymbol{\xi}_b) = s(\boldsymbol{k}; \boldsymbol{\xi}_a) + s(\boldsymbol{k}; \boldsymbol{\xi}_b) + b(\boldsymbol{k}). \tag{41}$$

The log posterior of the pair over the grid equals the log-likelihood of Eq. (8) up to an additive constant that is independent of the pair,

$$L_{a,b} = \sum_{\boldsymbol{k} \in \Omega} [V(\boldsymbol{k}) \ln v\,(\boldsymbol{k}; \boldsymbol{\xi}_a, \boldsymbol{\xi}_b) - v(\boldsymbol{k}; \boldsymbol{\xi}_a, \boldsymbol{\xi}_b)]\,. \tag{42}$$

The normalized posterior weight of the pair is

$$P_{a,b} = \frac{e^{L_{a,b}}}{\sum_{a'=1}^{S} \sum_{b'=1}^{S} e^{L_{a',b'}}}, \tag{43}$$

which sums to one over the $S^2$ ordered pairs and is exchangeable, i.e., $P_{a,b} = P_{b,a}$ by the symmetry of Eq. (41) in the two nodes.

### 4.2 The labeled MMSE estimator

The MMSE estimator of Eq. (17) is the posterior mean of each emitter index and is labeled because it is formed index by index, i.e., the first estimate is the posterior mean of the first emitter position and the second estimate is the posterior mean of the second, before any matching removes the labels. The marginal weight of the first index is

$$P_a^{(1)} = \sum_{b=1}^{S} P_{a,b} \tag{44}$$

and the first estimate is the corresponding weighted mean node

$$\hat{\boldsymbol{\theta}}_{\mathrm{MMSE},1} = \sum_{a=1}^{S} P_a^{(1)}\, \boldsymbol{\xi}_a. \tag{45}$$

The second marginal and the second estimate are formed in the same way from the second index. Since $P_{a,b} = P_{b,a}$, the two marginals are identical so the two estimates coincide, which is the degeneracy of Proposition 1. The computed separation $\left\|\hat{\boldsymbol{\theta}}_{\mathrm{MMSE},1} - \hat{\boldsymbol{\theta}}_{\mathrm{MMSE},2}\right\|$ is at the level of the floating point precision on every frame, so the degeneracy holds numerically and not only in the mean.

### 4.3 The matched Bayes estimator

The matched Bayes estimator of Eq. (28) minimizes the posterior mean of the matched loss $L_{\mathrm{M}}$ of Eq. (27). On the grid the objective of a candidate pair $(\boldsymbol{y}_1, \boldsymbol{y}_2)$ is

$$J(\boldsymbol{y}_1, \boldsymbol{y}_2) = \sum_{a=1}^{S}\sum_{b=1}^{S} P_{a,b} \min\left\{\|\boldsymbol{y}_i - \boldsymbol{\xi}_a\|^2 + \left\|\boldsymbol{y}_j - \boldsymbol{\xi}_b\right\|^2,\ i,j = 1,2, i \neq j\right\} \tag{46}$$

and the estimator is its minimizer

$$\hat{\boldsymbol{\theta}}_{\mathrm{B}} = \arg\min_{\boldsymbol{y}_1, \boldsymbol{y}_2} J(\boldsymbol{y}_1, \boldsymbol{y}_2). \tag{47}$$

The minimizer is found by a Lloyd iteration that alternates an assignment step and an update step. The assignment step matches every pair to the candidate by the smaller of the two terms of Eq. (46), which sets by the indicator function $\mathbf{1}(\cdot)$

$$I_{a,b} = \mathbf{1}\{\|\boldsymbol{y}_1 - \boldsymbol{\xi}_a\|^2 + \|\boldsymbol{y}_2 - \boldsymbol{\xi}_b\|^2 \leq \|\boldsymbol{y}_1 - \boldsymbol{\xi}_b\|^2 + \|\boldsymbol{y}_2 - \boldsymbol{\xi}_a\|^2\} \tag{48}$$

where $I_{a,b} = 1$ selects the identity match and $I_{a,b} = 0$ selects the swap. The update step sets each candidate to the posterior weighted mean of the nodes assigned to it,

$$\boldsymbol{y}_1 \leftarrow \sum_{a=1}^{S}\sum_{b=1}^{S} P_{a,b}\left[I_{a,b}\, \boldsymbol{\xi}_a + \left(1 - I_{a,b}\right) \boldsymbol{\xi}_b\right] \tag{49}$$

$$\boldsymbol{y}_2 \leftarrow \sum_{a=1}^{S}\sum_{b=1}^{S} P_{a,b}\left[I_{a,b}\, \boldsymbol{\xi}_b + \left(1 - I_{a,b}\right) \boldsymbol{\xi}_a\right] \tag{50}$$

where the weight of each candidate sums to one so no normalization is needed. The two steps do not increase $J$ at any iteration and converge to a fixed point. To keep the fixed point from being a local minimizer the iteration is started from the pair of largest posterior weight, i.e., the maximum a posteriori pair arg $\max_{(a,b)} P_{a,b}$, and from several random pairs, and the fixed point of smallest $J$ is kept as $\hat{\boldsymbol{\theta}}_{\mathrm{B}}$. By Eqs. (27) and (36) the objective $J$ equals the posterior mean of $MD_{\mathrm{H}}^2$ for $M = 2$ so $\hat{\boldsymbol{\theta}}_{\mathrm{B}}$ minimizes the reported metric by construction.

The iteration uses the posterior $p(\boldsymbol{\theta} \mid V)$of Eq. (12) and hence the true system parameters, so $\hat{\boldsymbol{\theta}}_{\mathrm{B}}$ computed this way uses the full system model at every frame, whereas the network of Section 3 approximates the same function of the frame in a single forward pass.

### *4.4 Grid resolution and Monte Carlo error*

The computation is exact up to two controlled errors. The first is the grid spacing $\Delta_g$ whose quantization standard deviation is, by the uniform distribution of an emitter position in the square of grid $\Delta_g$,

$$\sigma_g = \frac{\Delta_g}{\sqrt{12}}. \tag{51}$$

A spacing of $\Delta_g = 5$ to 6 nm gives $\sigma_g$ below 2 nm which is far smaller than the errors of the estimators compared in Section 5 so the grid does not affect the comparison. The second is the Monte Carlo average over the frames of one emitter separation. With $N = 200$ frames per separation, the relative standard error of the reported ARMSMD-H is near 5 percent and it falls as $N^{-1/2}$ so the full random dataset of $N = 1000$ frames reduces it to near 2 percent.

## 5. Simulation results

### *5.1 Simulation configuration*

**System parameters.** The camera has $7 \times 7$ pixels of size $\Delta_x = \Delta_y = 100$ nm, so the FOV is $700 \times 700$ nm$^2$, and the frame time is 0.01 s. The PSF is Gaussian with standard deviation 108.81 nm. Every emitter emits at $I = 3 \times 10^5$ photons per second, i.e., about 3000 photons per emitter per frame. The mixed noise is a frozen non-uniform field, i.e., the background density $b_k$ is a smooth spatial cloud on $[4, 6]$ photons/nm$^2$/s with a correlation length of 2 pixels while the readout density $G_k = \mu_k$ is drawn independently per pixel and uniformly on $[2, 4]$ with its mean set equal to its variance so the Poisson approximation of Section 2.1 applies. The field is generated once and saved, so every frame and every estimator consumes the identical $b_k$ and $G_k$ that produced the data. The prior draws the $M$ emitter positions independently and uniformly from the central $3 \times 3$ pixel block $\mathcal{B}$, which is the region on which the networks are trained.

**Testing datasets.** Two datasets are used in testing the trained networks and benchmarking by the EM-GML and UGIA-F estimators. First, Dataset-Random draws a fresh emitter configuration from the prior for each of $N = 1000$ frames per emitter number $M$, so it is drawn from the same distribution the networks are trained on and it is the in-distribution test. Second, Dataset-Fixed places the $M$ emitters equally spaced on a circle of radius $r$, so the nearest-neighbor separation is $2r\sin(\pi/M)$, which decreases as $M$ grows at fixed $r$. Each pair $(r, M)$ is repeated over $N = 25$ noise realizations with $r \in \{25, 50, 75, 100, 125, 150\}$ nm and $M \in \{1, \dots, 5\}$. The circle constellations are not drawn from the training prior, so Dataset-Fixed probes the generalization of the networks to configurations they were not trained on.

**Network parameters.** The five networks are sized to a common scale so that a result common to all of them reflects the inductive bias rather than the capacity. APL applies a linear map to two hidden layers of 256 and 128 units where every unit carries a learnable piecewise linear activation on 16 knots spaced uniformly on $[-3, 3]$. KAN uses two hidden layers of 64 and 32 units where every edge is a learnable univariate function that sums a sigmoid linear unit residual and a cubic B-spline on a grid of 5 intervals, i.e., a spline order of 3. CNN-APL and CNN-KAN prepend a two-layer convolutional stem of 16 channels with a $3 \times 3$ kernel and unit padding to the APL and KAN heads respectively, so the stem maps the $7 \times 7$ frame to 16 feature maps that flatten to $7 \times 7 \times 16 = 784$ values feeding the head, where the APL head carries two hidden layers of $256$ units and the KAN head carries hidden layers of $64$ and $32$ units. ViT reads each of the $49$ pixels as a token of a $128$ dimensional embedding, prepends a learnable class token and applies $6$ transformer encoder layers of $4$ self-attention heads each with a feedforward width of $512$ and a dropout of $0.1$, then maps the class token through a two-layer head of width $512$ to the output, which totals about $1.25$ million weights. Every network ends in a layer of $2M$ outputs read as $M$ positions in nanometers.

**Training.** All five networks are trained by minimizing the matched Hungarian loss $L_{\mathrm{M}}$ of Eq. (27) on frames generated on the fly from the prior, with the Adam optimizer and a cosine annealing learning rate over the run. Every network is trained with a batch size of $256$ for $40000$ steps, so each network sees $1.024 \times 10^7$ training frames, and a separate network is trained for each $M$. The learning rate is annealed from $10^{-3}$ to $10^{-5}$ for APL, CNN-APL, KAN and CNN-KAN and from $10^{-4}$ to $10^{-6}$ for ViT.

**Evaluation.** Every estimator is evaluated by the ARMSMD-H of Eq. (37), i.e., the root mean square minimum distance under the Hungarian assignment averaged over the frames. The UGIA-F and EM-GML estimators use the true system parameters at test time and, for UGIA-F, the true positions in addition, whereas the networks use the system parameters only in training to synthesize the frames and receive only the frame at test time. The EM-GML runs with multiple random initial positions in the support $\mathcal{B}$ to reach the global maximizer reliably, whereas a fully practical run initializes the EM by a rough estimator such as SIC [21] that uses no system information, so EM-GML is a practical estimator.

**Python code.** We developed the SMLM_Lib, a Python library for SMLM [38]. Based on it, custom Python code was developed for the simulations in this paper. The Python code can detect and take use of a GPU if it is available on a computer. All simulations were carried out on a Dell workstation XPS-8960 with Intel Core i7 of 2.10 GHz, 64 GB RAM and NVIDIA GeForce RTX 4070 of 12 GB VRAM.

The training times are listed in Table 1. The CNN head can speed up the convergence of APL and KAN in the training. Six example training frames are shown in Fig. 1.

**Table 1. Training time (min:sec) with 10.24 million training frames.**

| $M$ | APL | CNN-APL | KAN | CNN-KAN | ViT |
|---|---|---|---|---|---|
| 1 | 4:43 | 2:19 | 7:16 | 3:58 | 13.14 |
| 2 | 5:11 | 2:22 | 7:18 | 4:03 | 13:05 |
| 3 | 4:50 | 2:19 | 7:17 | 4:00 | 13:07 |
| 4 | 4:42 | 2:33 | 7:23 | 3:59 | 13:07 |
| 5 | 4:42 | 2:21 | 7:17 | 3:56 | 13:06 |

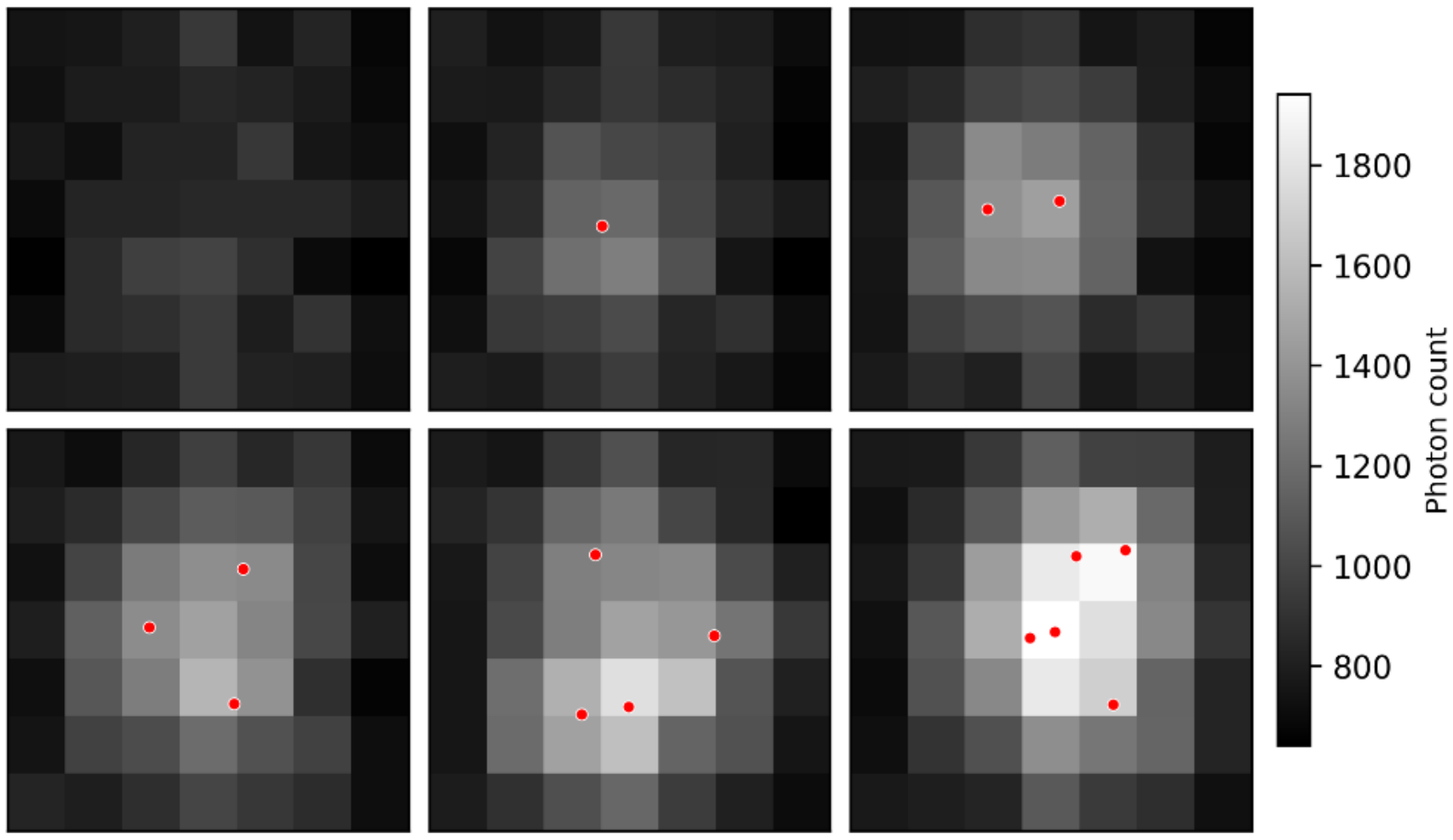

**Fig. 1.** Six example training frames. Top row $M = 0, 1, 2$ and bottom row $M = 3, 4, 5$, where $M = 0$ is a pure-noise frame showing the frozen non-uniform noise field. True emitter positions are shown as red dots. With the area of 0.09 μm$^2$ for the central $3 \times 3$ pixel block $\mathcal{B}$, the emitter density is 22.2, 33.3, 44.4, 55.6 emitters/μm$^2$ for $M = 2, 3, 4, 5$ respectively.

On the same workstation a trained network estimates one frame in about 0.5 to 16 μs at a saturating batch, i.e., between $6 \times 10^4$ and $2 \times 10^6$ frames per second or across the five networks. The multi-start EM of EM-GML at $K = 250$ starts and 300 iterations takes about 0.5 seconds per frame, so the networks are four to six orders of magnitude faster than the multi-start EM.

### 5.2 Two emitters against the Bayes optimum

For $M = 2$ the optimum matched Bayes estimator $\hat{\boldsymbol{\theta}}_{\mathrm{B}}$ and the MMSE estimator $\hat{\boldsymbol{\theta}}_{\mathrm{MMSE}}$ are computed by the quadrature of Section 4, so the estimators can be measured against the Bayes optimum on the same frames. Table 2 reports the accuracy and the agreement of every estimator on the 1000 frames of Dataset-Random, i.e., on frames drawn from the prior. The accuracy is the ARMSMD-H against the true positions, whereas the agreement is the ARMSMD-H against the Bayes estimate on the same frame, i.e., the direct measure of how closely an estimator reproduces $\hat{\boldsymbol{\theta}}_{\mathrm{B}}$ frame by frame.

**Table 2. Accuracy and Agreement with $\hat{\boldsymbol{\theta}}_{\mathrm{B}}$ in nm for $M = 2$.**

| Estimator | Bayes $\hat{\boldsymbol{\theta}}_{\mathrm{B}}$ | EM-GML | ViT | CNN-KAN | CNN-APL | KAN | APL | UGIA-F | MMSE |
|---|---|---|---|---|---|---|---|---|---|
| **Accuracy** | 13.21 | 14.03 | 14.37 | 14.53 | 14.90 | 16.60 | 17.18 | 17.19 | 85.97 |
| **Agreement** | 0 | 4.64 | 5.25 | 6.25 | 7.46 | 10.31 | 11.05 | 21.53 | 84.96 |

Three results follow from Table 2. First, no estimator falls below the Bayes optimum, i.e., $\hat{\boldsymbol{\theta}}_{\mathrm{B}}$ attains the smallest ARMSMD-H of 13.21 nm and a paired test over the frames confirms that no estimator is significantly below it, which validates the whole pipeline against the bound of Section 2.5. Second, every network is at or below UGIA-F, and the three stronger networks ViT, CNN-KAN and CNN-APL beat UGIA-F and approach both EM-GML and the Bayes optimum, with ViT within 1.2 nm of $\hat{\boldsymbol{\theta}}_{\mathrm{B}}$ and essentially equal to EM-GML. This holds although the networks receive only the frame at test time. Third, the MMSE estimator is degenerate as Proposition 1 predicts, i.e., its two estimates coincide to $7 \times 10^{-5}$ nm on every frame, which is the residual of the finite-precision quadrature rather than a nonzero separation, and its ARMSMD-H of 85.97 nm is that of the centroid.

The agreement row separates the networks from UGIA-F. EM-GML and ViT agree with the Bayes estimate to 4.64 and 5.25 nm, i.e., they reproduce $\hat{\boldsymbol{\theta}}_{\mathrm{B}}$ frame by frame and not only on average, whereas UGIA-F agrees only to 21.53 nm although its accuracy is similar to that of APL, i.e., it reaches a comparable average error by a structurally different route. The distinction is confirmed by the excess risk, i.e., the squared distance of an estimator from the true positions decomposes almost exactly into its squared distance from the Bayes estimate plus the squared distance of the Bayes estimate from the truth for the networks and for EM-GML. For example, for ViT, $14.37^2 \approx 5.25^2 + 13.21^2$. In contrast, the decomposition fails for UGIA-F. This supports reading the small excess of a network over the Bayes optimum as approximation error rather than a different kind of error.

The dependence on the emitter separation is shown in Fig. 2, i.e., the ARMSMD-H of every estimator against the separation $d$ at fixed $d$. UGIA-F diverges as the two emitters merge, i.e., it rises from about 10 nm at $d = 250$ nm to 116 nm at $d = 10$ nm as the Fisher information becomes near singular, whereas $\hat{\boldsymbol{\theta}}_{\mathrm{B}}$ and the networks remain bounded near 18 nm. The curves cross, i.e., UGIA-F is competitive only at large separation while the networks and $\hat{\boldsymbol{\theta}}_{\mathrm{B}}$ dominate at small separation. Fig. 2 conditions on a fixed separation, so it is a conditional view of the prior, and an estimator may fall below $\hat{\boldsymbol{\theta}}_{\mathrm{B}}$ at a single separation without contradicting the optimality of Section 2.5, which is an average over the prior. The optimality is the prior-averaged statement of Table 2, whereas Fig. 2 shows where in the configuration space each estimator gains or loses.

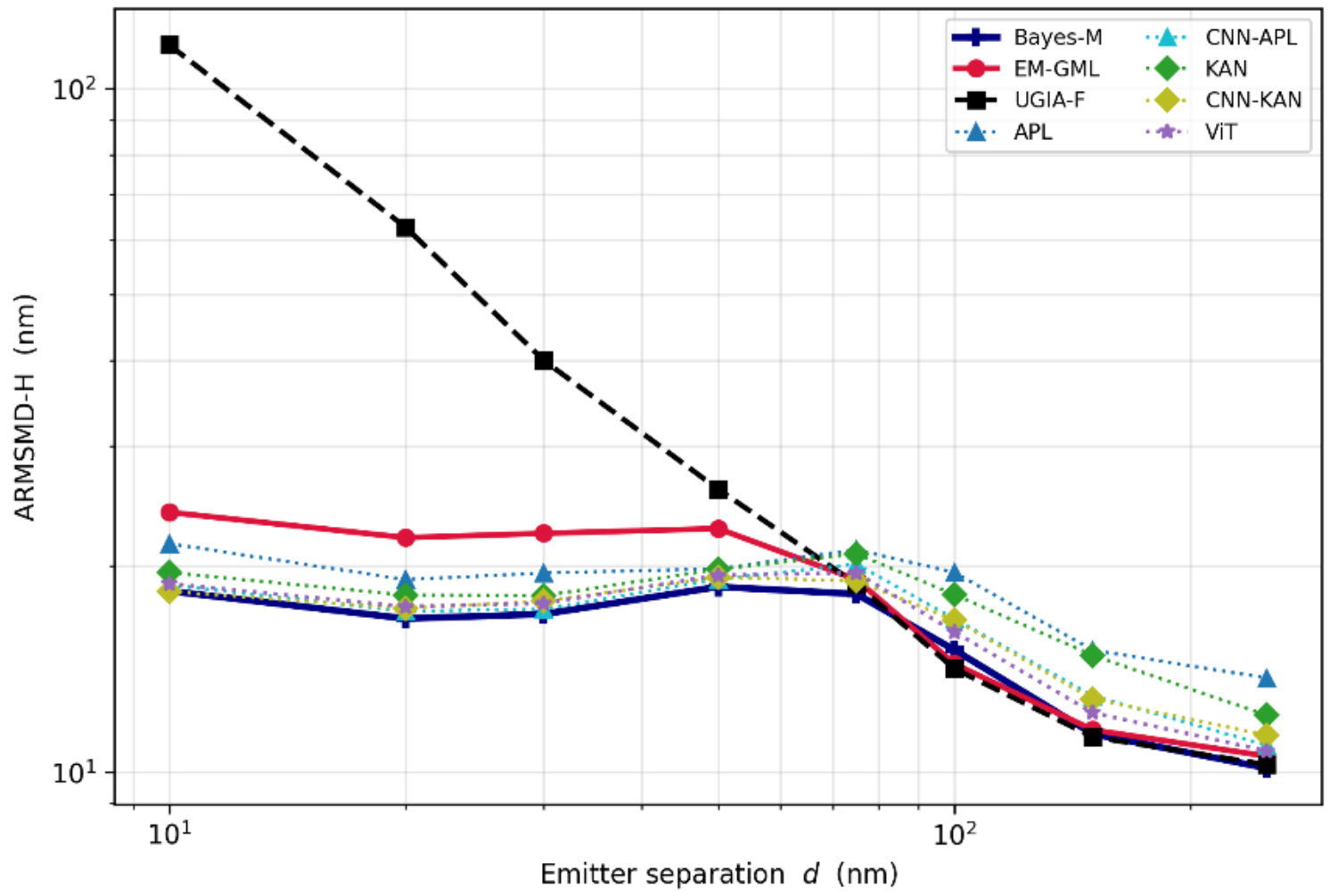

**Fig. 2.** Accuracy of estimators versus the Bayes optimum against the separation $d$ for $M = 2$ emitters.

### 5.3 Accuracy versus emitter number on the Random dataset

Table 3 reports the ARMSMD-H of every estimator on Dataset-Random for $M = 1$ to 5, i.e., the in-distribution accuracy as the number of emitters grows. With the area of 0.09 μm$^2$ for the central $3 \times 3$ pixel block $\mathcal{B}$ where the emitters are distributed, the emitter density is 22.2, 33.3, 44.4, 55.6 emitters/μm$^2$ for $M = 2, 3, 4, 5$ respectively, which are quite high. All estimators are evaluated on the same 1000 frames per $M$ and the UGIA-F Gaussian draw is seeded so the run is reproducible. Fig. 3 shows example images of estimated positions by the seven estimators for $M = 5$. More example images for $M = 1$ to 4 can be found in Ref. [39].

**Table 3. ARMSMD-H (nm) on Dataset-Random.**

| $M$ | APL | CNN-APL | KAN | CNN-KAN | ViT | UGIA-F | EM-GML |
|---|---|---|---|---|---|---|---|
| 1 | 9.63 | 9.34 | 9.34 | 9.35 | 9.77 | 9.46 | 9.49 |
| 2 | 17.22 | 14.90 | 16.60 | 14.53 | 14.37 | 25.80 | 14.02 |
| 3 | 29.53 | 20.71 | 24.63 | 22.56 | 21.27 | 49.65 | 20.61 |
| 4 | 36.30 | 28.02 | 32.11 | 28.55 | 27.26 | 116.35 | 26.11 |
| 5 | 37.73 | 31.67 | 35.64 | 33.22 | 31.59 | 172.39 | 30.30 |

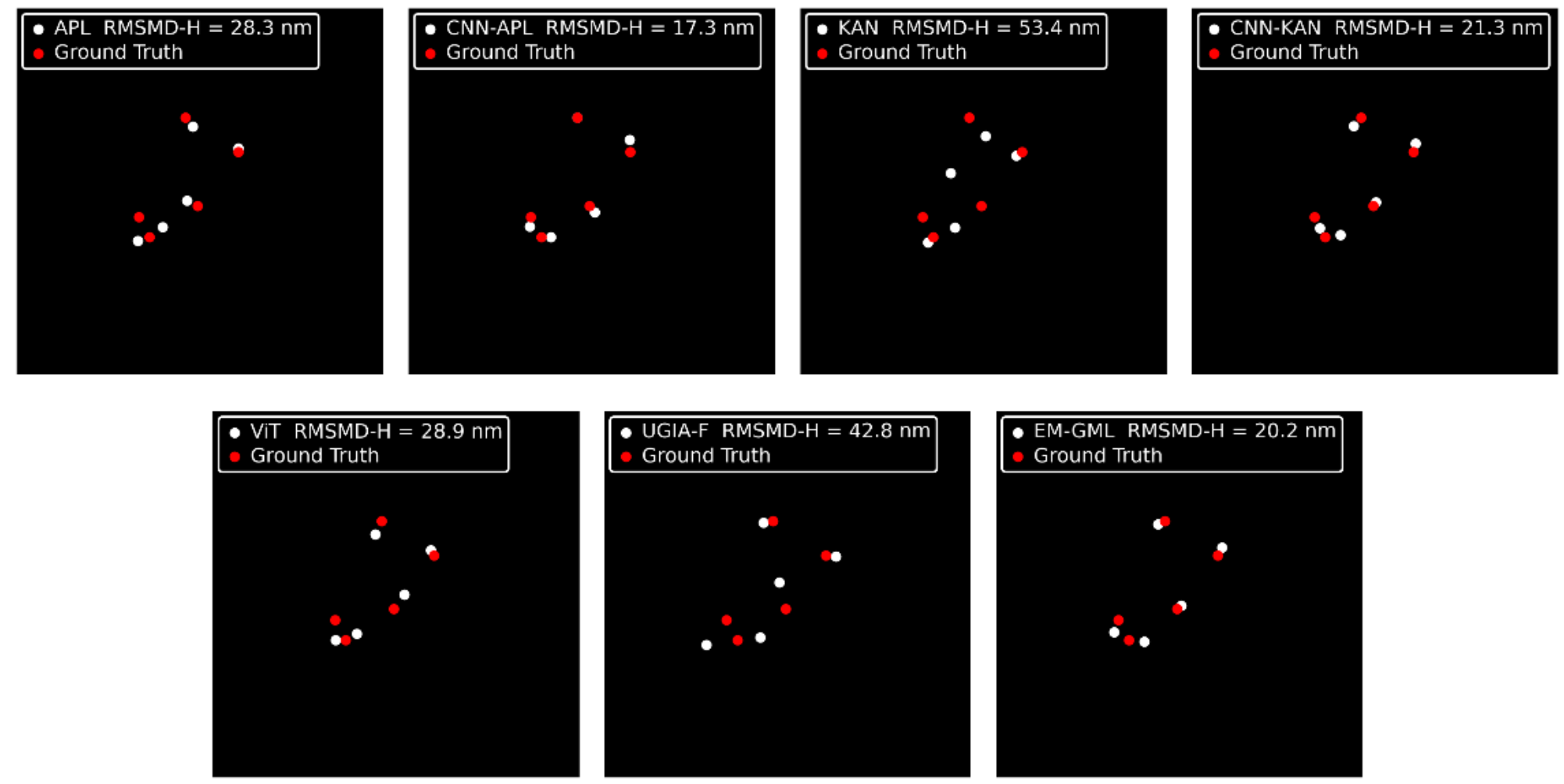

**Fig. 3.** Example images of positions estimated by the seven estimators for $M = 5$. The RMSMD-H for each image is shown.

For a single emitter there is no overlap and every estimator attains the same ARMSMD-H near 9.5 nm. As $M$ grows, the emitters overlap more often and UGIA-F degrades steeply, i.e., it rises to 172 nm at $M = 5$ because the CRB diverges whenever a pair of emitters merges. EM-GML remains the most accurate over the whole range and stays bounded, i.e., it grows only from 9.5 to 30 nm. Every network stays far below UGIA-F for $M \geq 3$ and approaches EM-GML, i.e., at $M = 5$ the best network ViT attains 31.6 nm against 30.3 nm for EM-GML and 172 nm for UGIA-F. The networks order themselves by architecture, i.e., the convolutional and attention networks ViT, CNN-KAN and CNN-APL lead while the plain APL and KAN trail, which shows that the advantage comes from learning from frames rather than from any single design. The hypothesis of the paper is therefore confirmed on the in-distribution test, i.e., the networks outperform the UGIA-F benchmark and approach EM-GML.

UGIA-F is realized once per frame from Eq. (34), i.e., a single Monte Carlo draw on the same footing as the single noisy frame that every other estimator receives, so it is reported from one realization rather than averaged. The aggregate is a high-variance quantity at overlap, i.e., it is dominated by the rare near-coincident frames where the CRB diverges. As sown in Table 5 of Section 5.5, for $M = 2$ the 17 frames with $d_{\min} < 25$ nm carry a conditional UGIA-F error of 162 nm, so a handful of frames lift the movie-wide value. The aggregate therefore varies between realizations, e.g., 25.8 nm here and 17.2 nm in the validation run of Table 2 on the same configurations, which is the sampling spread of the oracle rather than a discrepancy.

### *5.4 Generalization to the circle constellations*

Dataset-Fixed places the emitters on a circle of radius $r$, so the constellations are not drawn from the training prior and the nearest-neighbor separation $2r\sin(\pi/M)$ tightens as $M$ grows. This dataset therefore tests whether the networks generalize beyond the configurations they were trained on. The number of Monte Carlo runs is $N = 25$. Table 4 reports the ARMSMD-H of every estimator for each radius $r$ and emitter number $M$. Fig. 4 shows the estimated $N = 25$ positions by the seven estimators in comparison of the true positions for $r = 125$ nm and $M = 5$. All images of estimated positions by the seven estimators for $r = 25, 50, 75, 100, 125$ nm and $M = 1, 2, 3, 4, 5$ can be found in Ref. [39].

**Table 4. ARMSMD-H (nm) on Dataset-Fixed.**

| $r$ (nm) | $M$ | **APL** | **CNN-APL** | **KAN** | **CNN-KAN** | **ViT** | **UGIA-F** | **EM-GML** |
|---|---|---|---|---|---|---|---|---|
| 25 | 1 | 9.39 | 9.09 | 9.27 | 9.11 | 9.46 | 9.49 | 9.18 |
| 25 | 2 | 19.02 | 19.55 | 18.19 | 19.86 | 18.53 | 21.96 | 22.14 |
| 25 | 3 | 22.42 | 21.15 | 21.21 | 18.71 | 19.31 | 159.14 | 22.65 |
| 25 | 4 | 20.76 | 19.14 | 16.82 | 15.51 | 15.06 | 1625.92 | 23.77 |
| 25 | 5 | 23.55 | 19.86 | 15.44 | 19.93 | 13.62 | 280.29 | 21.34 |
| 50 | 1 | 6.85 | 6.58 | 6.74 | 6.85 | 6.93 | 10.79 | 6.77 |
| 50 | 2 | 17.89 | 17.28 | 15.90 | 15.61 | 15.43 | 13.50 | 15.54 |
| 50 | 3 | 33.38 | 33.95 | 30.88 | 30.10 | 30.74 | 49.12 | 30.93 |
| 50 | 4 | 26.78 | 29.14 | 27.42 | 28.18 | 27.40 | 185.83 | 31.78 |
| 50 | 5 | 21.53 | 26.21 | 20.87 | 29.01 | 22.34 | 1205.47 | 34.73 |
| 75 | 1 | 8.30 | 8.32 | 8.16 | 8.39 | 7.98 | 9.46 | 8.29 |
| 75 | 2 | 12.32 | 10.96 | 11.88 | 11.10 | 11.35 | 11.20 | 10.41 |
| 75 | 3 | 31.75 | 34.54 | 31.81 | 27.69 | 27.87 | 23.64 | 25.22 |
| 75 | 4 | 29.32 | 36.07 | 29.83 | 36.82 | 37.38 | 53.58 | 36.61 |
| 75 | 5 | 31.30 | 37.72 | 27.69 | 38.67 | 40.46 | 252.28 | 41.91 |
| 100 | 1 | 9.93 | 9.55 | 9.39 | 9.25 | 10.01 | 9.03 | 9.40 |
| 100 | 2 | 10.45 | 9.60 | 9.58 | 8.92 | 10.43 | 10.64 | 9.27 |
| 100 | 3 | 15.74 | 20.69 | 17.70 | 14.83 | 16.78 | 12.82 | 13.71 |
| 100 | 4 | 34.76 | 30.69 | 29.33 | 33.67 | 41.14 | 39.32 | 32.59 |
| 100 | 5 | 38.81 | 48.73 | 32.05 | 46.70 | 47.39 | 71.73 | 45.35 |
| 125 | 1 | 8.92 | 9.12 | 9.07 | 9.14 | 9.42 | 11.58 | 9.10 |
| 125 | 2 | 10.41 | 10.30 | 10.49 | 10.48 | 10.96 | 10.98 | 10.03 |
| 125 | 3 | 14.06 | 12.02 | 13.86 | 12.59 | 12.50 | 12.10 | 11.80 |
| 125 | 4 | 29.09 | 22.19 | 23.25 | 21.09 | 30.02 | 22.82 | 24.33 |
| 125 | 5 | 42.56 | 57.36 | 28.95 | 56.07 | 50.87 | 50.14 | 34.62 |
| 150 | 1 | 9.79 | 9.00 | 9.22 | 9.11 | 9.55 | 8.83 | 9.79 |
| 150 | 2 | 8.33 | 9.20 | 9.42 | 8.82 | 9.52 | 9.58 | 8.79 |
| 150 | 3 | 12.22 | 12.74 | 18.36 | 12.82 | 12.89 | 10.16 | 10.80 |
| 150 | 4 | 29.71 | 15.46 | 17.41 | 20.16 | 18.21 | 16.21 | 15.90 |
| 150 | 5 | 53.51 | 74.56 | 31.64 | 66.98 | 57.15 | 27.20 | 35.32 |

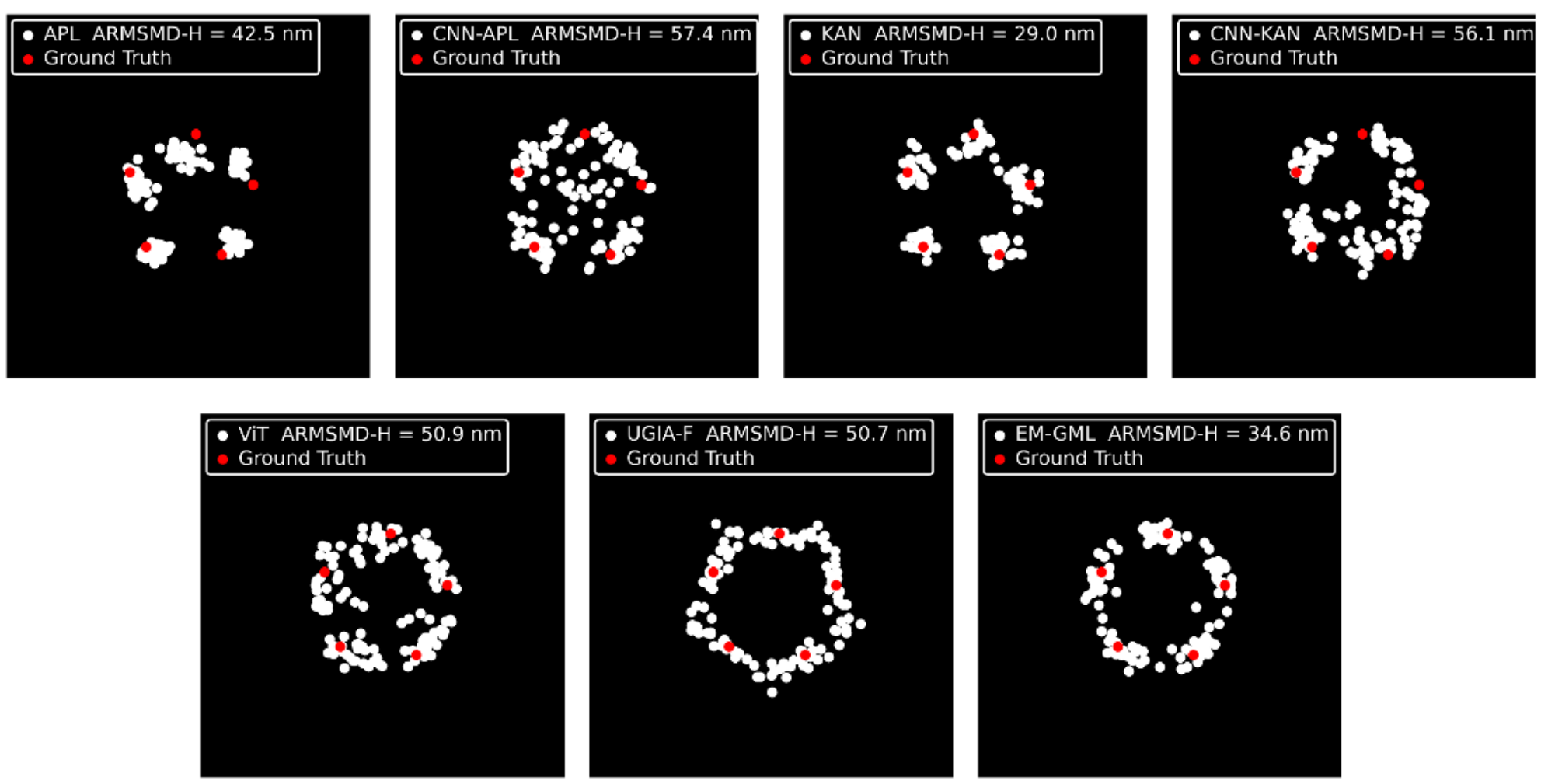

**Fig. 4.** The images of positions estimated by the seven estimators over $N = 25$ frames for a radius of $r = 125$ nm and $M = 5$ emitters.

The dominant feature of Table 4 is the divergence of UGIA-F at severe overlap, i.e., the smallest radius with the largest emitter number packs the circle within the PSF so the Fisher information becomes near singular and the best unbiased accuracy is unbounded, e.g., UGIA-F reaches 1626 nm at $r = 25$ nm and $M = 4$, and 1205 nm at $r = 50$ nm and $M = 5$, whereas no learned or likelihood estimator exceeds 75 nm anywhere in the table. This is the CRB divergence of Section 2.7 made concrete, i.e., the biased EM-GML and the networks remain bounded where the oracle UGIA-F does not. At large radius all estimators converge to the single-emitter ARMSMD-H near 9 to 15 nm, i.e., the overlap is resolved and the estimators agree.

Between these limits the networks generalize well but not perfectly. They track EM-GML across most of the table, whereas at the tightest packings and the largest $M$ some networks exceed EM-GML, e.g., CNN-APL reaches 74.6 nm at $r = 150$ nm and $M = 5$. This is consistent with Section 3.1, i.e., the network approximation of the Bayes estimator holds on the training prior and degrades on constellations far from it, so Dataset-Fixed marks the edge of generalization while Dataset-Random of Section 5.3 is the fair in-distribution comparison.

### *5.5 Stratified analysis by emitter overlap*

The aggregate of Section 5.3 averages over all overlap severities, so it hides where in the configuration space a network overtakes a benchmark. The per-frame results of Dataset-Random are therefore stratified by the minimum emitter separation $d_{\min}$ of each frame, which isolates the effect of overlap from the effect of train-test mismatch because every stratum is drawn from the training prior. Table 5 reports the conditional ARMSMD-H within each stratum for $M = 2, 3, 4, 5$. The case $M = 1$ is excluded because a single emitter has no pairwise separation and $d_{\min}$ is undefined.

**Table 5. Conditional ARMSMD-H on Dataset-Random by minimum emitter separation $d_{\min}$.**

| $M$ | $d_{min}$ (nm) | $n$ | APL | CNN-APL | KAN | CNN-KAN | ViT | UGIA-F | EM-GML |
|---|---|---|---|---|---|---|---|---|---|
| 2 | [0, 25) | 17 | **22.71** | **20.19** | **19.87** | **20.13** | **20.78** | 161.89 | 26.44 |
| 2 | [25, 50) | 53 | **18.19** | **15.69** | **16.24** | **16.88** | **16.15** | 29.07 | 20.80 |
| 2 | [50, 75) | 69 | 19.11 | **17.82** | 19.25 | 18.59 | 19.24 | 24.34 | 18.53 |

| 2 | [75, 100) | 114 | 21.05 | 17.60 | 19.82 | 18.49 | 17.85 | 17.80 | 16.75 |
|---|---|---|---|---|---|---|---|---|---|
| 2 | [100, 150) | 231 | 18.99 | 15.35 | 18.14 | 14.81 | 15.43 | 12.77 | 13.68 |
| 2 | ⩾150 | 516 | 14.68 | 13.22 | 14.51 | 12.11 | 11.51 | 10.85 | 11.03 |
| 3 | [0, 25) | 71 | 29.28 | **19.83** | 25.92 | 24.95 | **21.28** | 156.39 | 23.16 |
| 3 | [25, 50) | 180 | 27.33 | **20.48** | 23.22 | **21.60** | **21.03** | 47.38 | 22.50 |
| 3 | [50, 75) | 182 | 28.06 | **20.89** | 23.97 | **21.34** | **21.58** | 30.31 | 23.38 |
| 3 | [75, 100) | 191 | 28.39 | 22.29 | 27.09 | 25.01 | 22.63 | 21.17 | 21.40 |
| 3 | [100, 150) | 261 | 32.73 | 20.80 | 26.12 | 23.72 | 22.20 | 14.35 | 18.54 |
| 3 | ⩾150 | 115 | 29.45 | 18.19 | 18.42 | 16.55 | 15.99 | 12.39 | 12.86 |
| 4 | [0, 25) | 116 | 33.31 | 26.93 | 28.94 | **24.73** | **24.07** | 307.99 | 25.86 |
| 4 | [25, 50) | 264 | 36.25 | 27.38 | 32.65 | 28.18 | 26.87 | 85.46 | 25.71 |
| 4 | [50, 75) | 263 | 36.15 | 27.95 | 31.92 | 28.62 | **26.95** | 36.39 | 27.69 |
| 4 | [75, 100) | 214 | 36.09 | 28.37 | 31.73 | 29.50 | 28.54 | 30.49 | 26.35 |
| 4 | [100, 150) | 134 | 39.50 | 30.18 | 34.75 | 30.83 | 29.55 | 20.10 | 23.95 |
| 4 | ⩾150 | 9 | 34.37 | 19.26 | 28.23 | 24.23 | 17.61 | 19.98 | 16.37 |
| 5 | [0, 25) | 199 | 36.95 | 29.81 | 34.17 | 31.76 | 30.16 | 354.44 | 29.05 |
| 5 | [25, 50) | 349 | 36.98 | 31.59 | 34.67 | 31.42 | 29.99 | 104.74 | 29.44 |
| 5 | [50, 75) | 298 | 37.84 | 31.15 | 36.04 | 33.18 | 32.62 | 48.29 | 29.96 |
| 5 | [75, 100) | 123 | 39.33 | **34.01** | 38.04 | 37.34 | **34.05** | 37.88 | 34.20 |
| 5 | [100, 150) | 31 | 43.15 | 38.76 | 41.57 | 43.43 | 37.50 | 24.86 | 34.19 |

Note: The count $n$ is the number of frames in the stratum. A network value is set in bold where it is smaller than the EM-GML value in the same stratum and the same $M$, whereas the oracle UGIA-F is never marked.

Two patterns stand out. First, the advantage of the networks over the oracle UGIA-F is concentrated in the overlapped strata, i.e., at the smallest $d_{\min}$ the networks are far more accurate than the diverging UGIA-F, e.g., for $M = 2$ the best network reaches $19.9$ nm against $162$ nm for UGIA-F in the $[0, 25)$ stratum, whereas at $d_{\min} \geq 150$ nm the oracle is best and the networks trail by a few nanometers. Every network is more accurate than UGIA-F throughout the overlapped strata below $75$ nm at all $M$. Second, the networks also beat EM-GML in the most overlapped strata at low emitter number, which the bold entries of Table 5 mark, i.e., for $M = 2$ every network is more accurate than EM-GML in the two strata below $50$ nm, whereas for $M = 5$ EM-GML leads in every stratum except that ViT and CNN-APL edge past it in the $[75, 100)$ stratum.

### *5.6 Verification of the EM-GML benchmark*

EM-GML serves as the maximum likelihood benchmark in every table, so the multi-start EM that realizes it must be shown to reach the global maximum of the likelihood rather than a local one. For $M = 2$ the parameter space of emitter positions is four dimensional and can be searched exhaustively, which provides a reference that the multi-start is measured against. An exhaustive grid search over the whole parameter space at a grid step of 5 nm gives the maximizing pair of grid positions on each frame, which is then refined by EM to the exact off-grid maximum of the mode it occupies and is taken as the reference global maximum. On a frame the multi-start is counted as reaching the reference when its largest log-likelihood over the $K$ starts is within a tolerance of the reference log-likelihood. The comparison is on the log-likelihood rather than on the position because two configurations far apart in position can be nearly tied in likelihood when the surface is flat.

Fig. 5 shows the fraction of the $1000$ frames on which the multi-start reaches the reference and the mean log-likelihood shortfall below it against the number of starts $K$. The fraction rises from $66$ percent at $K = 1$ to $99.7$ percent at $K = 250$ whereas the mean shortfall falls by more than three orders of magnitude to about $10^{-5}$, i.e., even on the few frames that do not reach the reference exactly the remaining gap is negligible. A separate check finds that raising the EM iteration budget from $300$ to $1000$ changes the fraction little, so the budget of $300$ iterations used throughout is sufficient. The multi-start EM with $K = 250$ therefore attains the global maximum likelihood estimate reliably for $M = 2$, which justifies the use of EM-GML as the GML benchmark in this paper.

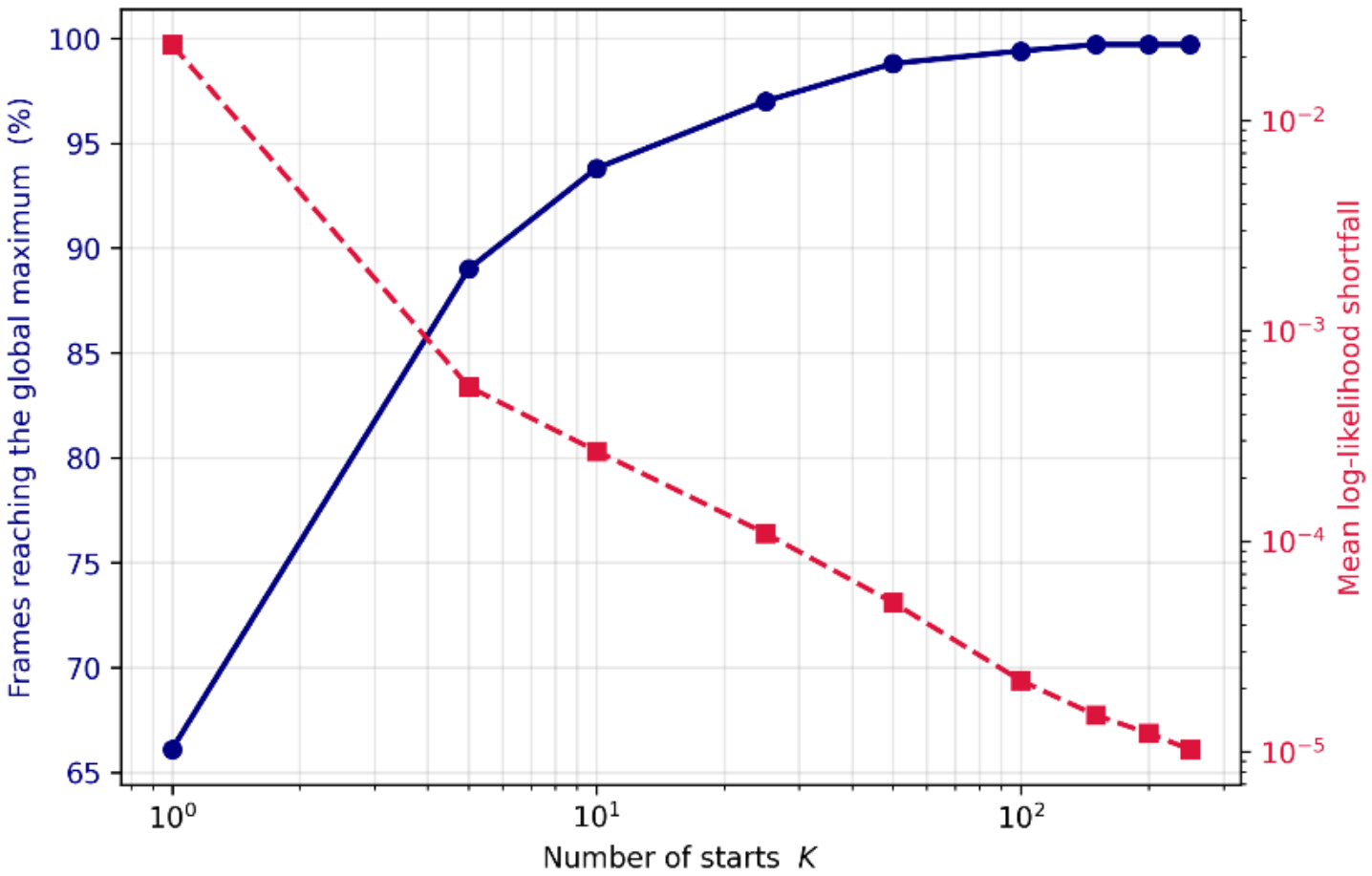

**Fig. 5.** Verification of the EM-GML benchmark against an exhaustive grid search for $M = 2$. The navy curve is the percentage of the 1000 Dataset-Random frames on which multi-start EM reaches the reference global maximum, whereas the crimson curve is the mean log-likelihood shortfall below the reference, both against the number of EM starts $K$.

## 6. Discussion

### *6.1 Information used by the three estimators and their practical usefulness*

The three estimators differ in the information they use when they estimate the emitter positions from a given frame. A network uses only the frame, since the system parameters and the true positions enter only in training through the synthesized frames and their labels. EM-GML uses the system parameters, i.e., it maximizes the per-frame likelihood and does not use the true positions. UGIA-F uses the system parameters and the true positions, since its Gaussian draw of Eq. (34) is centered at the true positions. Only UGIA-F therefore requires information that a real experiment cannot supply, so UGIA-F alone is an oracle whereas the networks and EM-GML are practically useful estimators.

At estimation time the networks use the least, i.e., the frame alone, and estimate by a single non-iterative forward pass that is much faster than the iterative EM of EM-GML as reported in Section 5.1, so once trained they are the most readily deployed. EM-GML needs the calibrated system model of every frame but not the position truth and is deployable wherever the model is known. UGIA-F cannot be run on data whose positions are unknown and serves only as the benchmark of the best accuracy attainable by an unbiased estimator.

The comparison is nonetheless made on the same system model, i.e., EM-GML and UGIA-F use it directly while the networks embed it in their training frames, and the networks additionally use the prior over configurations. A frame-trained network converts the observed frame into an estimate through the fixed map learned in training, which encodes both the system model of the synthesized frames and the prior, whereas EM-GML maximizes the per-frame likelihood under a flat prior. The prior is the side information that lets a biased estimator stay bounded where the unbiased CRB diverges, so the networks and EM-GML remain accurate at overlap while UGIA-F does not, and the networks are free to match EM-GML and to outperform UGIA-F on the frames drawn from the prior they were trained on.

This is why the results of Section 5.3 and Section 5.5 are not paradoxical. The networks beat the oracle UGIA-F over the overlapped strata not by using more of the system but by carrying the prior, which UGIA-F does not use, so a biased learned estimator stays accurate wherever the prior is informative about the overlapped configurations, which is exactly where the unbiased UGIA-F is weakest.

### 6.2 The network as a practical approximation of the Bayes estimator

Section 2.5 and Section 4 establish that the matched Bayes estimator $\hat{\boldsymbol{\theta}}_{\mathrm{B}}$ has no closed form for $M \geq 2$ and is computed by a fixed-point iteration that uses the posterior of Eq. (12) and hence the full system model. A direct computation of $\hat{\boldsymbol{\theta}}_{\mathrm{B}}$ therefore uses the posterior and hence the full system model at every frame, i.e., the same per-frame model computation that EM-GML performs, though unlike UGIA-F it does not need the true positions. The network amortizes this computation, i.e., it is trained to approximate the same function of the frame that $\hat{\boldsymbol{\theta}}_{\mathrm{B}}$ computes and, once trained, evaluates it in a single forward pass without the per-frame posterior integration. The forward pass is non-iterative, so once trained the network estimates far faster than the iterative EM of EM-GML and the iterative fixed point that computes $\hat{\boldsymbol{\theta}}_{\mathrm{B}}$, which is an important practical advantage of the network.

The simulation supports this reading in two ways. On Dataset-Random of Section 5.3 the networks track EM-GML and beat the oracle UGIA-F over the overlapped strata, which is the behavior expected of an estimator that minimizes the mean square error over the prior. On Dataset-Fixed of Section 5.4 the accuracy degrades on the tightest circle constellations, i.e., exactly the configurations that lie farthest from the prior, which confirms that the network approximates $\hat{\boldsymbol{\theta}}_{\mathrm{B}}$ on the training prior rather than a parameter-free estimator that would hold everywhere. The scope of the approximation is therefore the prior, and within that scope the network reaches the accuracy of EM-GML, which is the hypothesis of the paper.

Hence, the results present a strong promise of neural networks in developing high-throughput large-FOV super spatiotemporal resolution SMLM [40].

### 6.3 Necessity of the Hungarian assignment

Sections 2.3 to 2.5 theoretically analyze the performance of the MMSE estimator, the sorted Bayes estimator, and the matched Bayes estimator. Correspondingly, their loss functions require no match, one coordinate match, and full match between the estimated position and the true positions. In simulations, Section 5.2 reports the accuracy of the MMSE estimator that generates a single estimated position for both of the $M = 2$ emitters, confirming the theoretical result in Section 2.3 that the estimated position coordinates in both $x$ and $y$ axes merge to a single position. In addition, the networks in all reported simulation results are trained by using the Hungarian assignment aiming to approach the matched Bayes estimator. Their reconstructed images of estimated emitter positions show balanced small spread over both $x$ and $y$ coordinates, confirming the theoretical analysis in Section 2.5.

Though not reported in detail, we also trained the five networks to approach the sorted Bayes estimator by matching only $x$ coordinates between the estimated and the true positions. The result clearly shows that when the $x$ coordinates of two emitter positions are near, their network estimated $y$ coordinates are widely spread, confirming the theoretical analysis in Section 2.4.

Hence, for a neural network to approach the *optimum* matched Bayes estimator, the Hungarian assignment must be applied in training.

### 6.4 Why the oracle UGIA-F diverges while the biased estimators stay bounded

The most visible feature of Table 3 and Table 4 is the divergence of UGIA-F at severe overlap against the bounded error of EM-GML and the networks. This follows from the status of unbiasedness as a constraint rather than an optimality property, which is stated in Section 2.7. When a pair of emitters merges, the Fisher information matrix becomes near singular and the CRB diverges, so the best accuracy attainable under the unbiasedness constraint is unbounded and UGIA-F inherits that divergence through Eq. (34). EM-GML carries a bias and is therefore free of the constraint, so its error stays bounded where the CRB does not, e.g., it grows only from $9.5$ to $30$ nm across $M = 1$ to $5$ on Dataset-Random whereas UGIA-F reaches $172$ nm.

The networks share the bounded behavior of EM-GML rather than the divergence of UGIA-F. A network minimizes the matched loss of Eq. (27) averaged over the prior, i.e., it minimizes a mean square error and not a variance under an unbiasedness constraint, so it too is a biased estimator and is not bounded by the CRB. The prior supplies the information that a biased estimator needs to stay accurate when the frame alone is nearly uninformative, i.e., when two emitters overlap the prior still favors the configurations it was trained on and the network settles on a plausible pair rather than on the diverging unbiased solution. The divergence of UGIA-F and the boundedness of the learned estimators are therefore two sides of the same fact, i.e., an unbiased estimator pays for the merging of the emitters through the CRB whereas a biased estimator does not.

## 7. Conclusion

This paper tested the hypothesis that a neural network trained on synthesized frames can outperform the unbiased oracle UGIA-F and approach the likelihood estimator EM-GML in multi-emitter localization. Five networks were studied, i.e., APL, KAN, CNN-APL, CNN-KAN and ViT, each trained on frames drawn on the fly from the prior and each receiving only the frame at test time, whereas UGIA-F and EM-GML receive the PSF, the emitter intensity and the per-pixel noise densities, and UGIA-F receives the true positions as well. The matched Bayes estimator was identified as the target that a frame-trained network approximates, and Section 4 verified by quadrature on two emitters that this estimator has no closed form and is computed by a fixed-point iteration that needs the full system model, so a direct computation itself uses the full system model at every frame.

The simulation confirmed the hypothesis on the in-distribution test. On Dataset-Random every network stayed far below UGIA-F for $M \geq 3$ and approached EM-GML, e.g., at $M = 5$ the best network ViT attained $31.6$ nm against $30.3$ nm for EM-GML and $172$ nm for UGIA-F. The networks ordered themselves by architecture while all of them beat the oracle UGIA-F, which shows that the advantage comes from learning from frames rather than from any single design. The stratified analysis located the advantage in the overlapped configurations, i.e., where the emitters merge, the networks were far more accurate than the diverging UGIA-F and, at low emitter number, more accurate than EM-GML as well.

Two findings frame the result. First, the divergence of the oracle UGIA-F at overlap and the bounded error of the networks are two sides of unbiasedness being a constraint rather than an optimality property, i.e., a biased estimator is free of the CRB that the merging emitters make unbounded. Second, at estimation time the networks use only the frame, EM-GML uses the system parameters, and UGIA-F uses the system parameters and the true positions, so only UGIA-F is an oracle while the networks and EM-GML are practically useful, and the networks additionally use the prior in training, so their accuracy comes from learning the Bayes-optimal estimator rather than from any information advantage.

The approximation holds on the prior it was trained on. On Dataset-Fixed the accuracy degraded on the tightest circle constellations, i.e., the configurations that lie farthest from the training prior, which marks the edge of generalization and confirms that the network approximates the matched Bayes estimator on the prior rather than a parameter-free estimator that would hold everywhere. Within that scope the networks reached the accuracy of EM-GML, so a frame-trained network is a practical estimator for multi-emitter localization, i.e., once trained on frames synthesized for the imaging conditions of interest it evaluates the Bayes-optimal estimate in a single non-iterative forward pass, which is much faster than the iterative EM that EM-GML requires, presenting a strong promise of neural networks in high-throughput large-FOV SMLM.

This paper is a proof of concept. The capability of a network to approach the optimal matched Bayes estimator, i.e., the estimator that minimizes the mean square error over the prior, is established here in concept by simulation on a FOV of $700 \times 700$ nm$^2$. With the emitters distributed in the central $3 \times 3$ pixel block of area $0.09$ μm$^2$, the corresponding emitter density of 22.2, 33.3, 44.4, 55.6 emitters/μm$^2$ for $M = 2, 3, 4, 5$ is quite high. This study suggests that supported with an advanced GPU, the convolutional and attention networks CNN-APL, CNN-KAN and ViT can be trained on frames of a practically large size, e.g., $2048 \times 2048$ pixels, to approach the optimum Bayes estimator to eventually realize high-throughput large-FOV super spatiotemporal resolution SMLM.

## Appendix A: Proofs

### *A.1 Proof of Proposition 1*

By Eqs. (2), (3), (5) and (6) the frame mean $v(\boldsymbol{k})$ depends on $\boldsymbol{\theta}$ only through the sum $Q(\boldsymbol{k})$, which is symmetric in the emitter index, so $f_V(V; \mathbf{P}_\pi \boldsymbol{\theta}) = f_V(V; \boldsymbol{\theta})$ for every $\pi \in \mathcal{S}_M$. Together with the exchangeability of the prior and Eq. (12) this gives $p(\mathbf{P}_\pi \boldsymbol{\theta} \mid V) = p(\boldsymbol{\theta} \mid V)$, so the posterior is exchangeable and all its marginals are identical. The conditional means are therefore equal. ■

### *A.2 Proof of Corollary 1*

Write $\hat{\boldsymbol{m}} = \boldsymbol{c} + \boldsymbol{e}$ and note that $\boldsymbol{c} - \boldsymbol{\theta}_1 = -(\boldsymbol{c} - \boldsymbol{\theta}_2)$ and $\|\boldsymbol{c} - \boldsymbol{\theta}_m\| = d/2$ for $m = 1,2$. Expanding both squared norms and summing, the cross terms cancel and Eq. (19) follows. ■

### *A.3 Proof of Proposition 2*

Take $x_1 \leq x_2$ without loss of generality and let $A = \{X_1 < X_2\}$ be the event that the order is preserved. The order is reversed exactly when $X_1 > X_2$, so by Eq. (22) $\mathrm{P}(A^{\mathrm{c}}) = p$ and $\mathrm{P}(A) = 1 - p$. The indicator $\mathbf{1}_A$ depends only on $\xi_1$ and $\xi_2$ and is therefore independent of $Y_1$ and $Y_2$.

Consider first $Y_{(1)}$. Since $Y_{(1)} = \mathbf{1}_A Y_1 + (1 - \mathbf{1}_A) Y_2$ with $\mathbf{1}_A$ independent of $(Y_1, Y_2)$,

$$\mathbb{E}\left[Y_{(1)}\right] = (1 - p)\, y_1 + p\, y_2, \tag{52}$$

$$\mathbb{E}\left[Y_{(1)}^2\right] = \sigma^2 + (1 - p)\, y_1^2 + p\, y_2^2. \tag{53}$$

Subtracting the square of the mean in Eq. (52) from Eq. (53),

$$\mathrm{Var}\left(Y_{(1)}\right) = \sigma^2 + p(1 - p)(y_1 - y_2)^2, \tag{54}$$

which is Eq. (23).

Consider next $X_{(1)} = \min(X_1, X_2) = S - |T|/2$ with $S = (X_1 + X_2)/2$ and $T = X_2 - X_1$. Since $X_1$ and $X_2$ are independent with the common variance $\sigma^2$, $S$ and $T$ are uncorrelated and jointly Gaussian, hence independent, with $\mathrm{Var}(S) = \sigma^2/2$ and $T \sim \mathcal{N}(d_x, 2\sigma^2)$. Therefore

$$\begin{aligned}\mathrm{Var}\left(X_{(1)}\right) &= \frac{\sigma^2}{2} + \frac{1}{4}\left[\mathbb{E}(T^2) - \mathbb{E}^2(|T|)\right] \\ &= \sigma^2 + \frac{1}{4}\left[d_x^2 - \mathbb{E}^2(|T|)\right],\end{aligned} \tag{55}$$

using $\mathbb{E}[T^2] = 2\sigma^2 + d_x^2$. The first absolute moment of the folded normal has the closed form

$$\mathbb{E}(|T|) = \frac{2\sigma}{\sqrt{\pi}} \exp\left(-\frac{d_x^2}{4\sigma^2}\right) + d_x\,(1 - 2p), \tag{56}$$

where $\tau^2 = 2\sigma^2$ and $2\Phi(d_x/\tau) - 1 = 1 - 2p$ by Eq. (22). Since $T = X_2 - X_1$, Eq. (56) is Eq. (24), and substituting it into Eq. (55) gives Eq. (25).

The two limits follow by inspection. At $d_x = 0$ the exponential equals one and $1 - 2p = 0$, so $\mathbb{E}(|T|) = 2\sigma/\sqrt{\pi}$ and $\mathrm{Var}(X_{(1)}) = \sigma^2(1 - 1/\pi)$; as $d_x \to \infty$ the exponential vanishes and $1 - 2p \to 1$, so $\mathbb{E}(|T|) \to d_x$ and $\mathrm{Var}(X_{(1)}) \to \sigma^2$. Moreover $\mathrm{Var}(X_{(1)})$ is nondecreasing in $d_x$, since $\mathrm{d}\,\mathbb{E}(|T|)/\mathrm{d}\,d_x = 1 - 2p$ and hence $\mathrm{d}\,\mathrm{Var}(X_{(1)})/\mathrm{d}\,d_x = \frac{1}{2}[d_x - \mathbb{E}(|T|)\,(1 - 2p)] = \frac{1}{2}\,\mathrm{Cov}(|T|, \mathrm{sign}(T)) \geq 0$, where the covariance is nonnegative because $\mathbb{E}[\mathrm{sign}(T) \mid |T| = r] = \tanh(d_x r/\tau^2)$ is nondecreasing in $r$. Hence $\sigma^2(1 - 1/\pi) \leq \mathrm{Var}(X_{(1)}) \leq \sigma^2$. ■

### *A.4 Proof of Corollary 2*

By Eqs. (14) and (15) the estimator $\hat{\boldsymbol{\theta}}_{\mathrm{B}}$ minimizes the inner integral of Eq. (14) with $L = L_{\mathrm{M}}$ for every $V$, so it minimizes $R$ under $L_{\mathrm{M}}$ over all functions of the frame. Both $\hat{\boldsymbol{\theta}}_{\mathrm{S}}$ and $\hat{\boldsymbol{\theta}}_{\mathrm{MMSE}}$ are functions of the frame. ■

### Funding

### Acknowledgments

### Disclosures

The authors declare no conflicts of interest.

### Data Availability

Data underlying the results presented in this paper, i.e., the paper draft and all the Python scripts built on SMLM_Lib that generate the datasets, train the networks and produce the tables, are available in Ref. [39]. The SMLM_Lib library is available in Ref. [38].